\documentclass[journal]{IEEEtran}

\ifCLASSINFOpdf
\else
   \usepackage[dvips]{graphicx}
\fi
\usepackage{url}
\usepackage{graphicx}
\usepackage{cite}
\usepackage{amsmath,amssymb,amsfonts}

\usepackage{textcomp}
\usepackage{subfigure}
\usepackage{caption}
\usepackage{xcolor}
\usepackage{booktabs}
\usepackage{arydshln}
\usepackage{bbding}
\usepackage{multirow}
\usepackage{diagbox}
\usepackage{ntheorem}
\usepackage{makecell}
\usepackage{colortbl}
\usepackage{pifont}% http://ctan.org/pkg/pifont
\usepackage{tablefootnote}

\usepackage[switch]{lineno} 
\usepackage{siunitx}
\usepackage[pagebackref=true,breaklinks=true,colorlinks,bookmarks=false]{hyperref}

\usepackage[font={bf},textfont=md]{caption}

\newcommand{\figref}[1]{Fig.~\ref{#1}}
\newcommand{\equref}[1]{Eq.~\eqref{#1}}
\newcommand{\secref}[1]{Sec.~\ref{#1}}
\newcommand{\tabref}[1]{Table.~\ref{#1}}

\usepackage{xspace}
\makeatletter
\DeclareRobustCommand\onedot{\futurelet\@let@token\@onedot}
\def\@onedot{\ifx\@let@token.\else.\null\fi\xspace}

\def\eg{\emph{e.g}\onedot} 
\def\ie{\emph{i.e}\onedot} 
\def\cf{\emph{c.f}\onedot} 
\def\etc{\emph{etc}\onedot} 
 
\def\etal{\emph{et al}\onedot}
\makeatother

\definecolor{seagreen}{RGB}{84,255,159}
\definecolor{SpringGreen}{RGB}{0,139,69}

\newcolumntype{C}[1]{>{\PreserveBackslash\centering}p{#1}}
\newcolumntype{R}[1]{>{\PreserveBackslash\raggedleft}p{#1}}
\newcolumntype{L}[1]{>{\PreserveBackslash\raggedright}p{#1}}

\begin{document}

\title{VRNet: Learning the Rectified Virtual Corresponding Points for 3D Point Cloud Registration}

\author{Zhiyuan Zhang, Jiadai Sun, Yuchao Dai \IEEEmembership{Member, IEEE}, Bin Fan, and Mingyi He \IEEEmembership{Senior Member, IEEE}
\thanks{Zhiyuan Zhang, Jiadai Sun, Yuchao Dai, Bin Fan and Mingyi He are with School of Electronics and Information, Northwestern Polytechnical University, Xi'an 710129, China (email:\{zhangzhiyuan, sunjiadai, binfan\}@mail.nwpu.edu.cn, daiyuchao@gmail.com, myhe@nwpu.edu.cn). Yuchao Dai is the corresponding author.}
}

\markboth{Journal of \LaTeX\ Class Files, Vol. 14, No. 8, August 2015}
{Shell \MakeLowercase{\textit{et al.}}: Bare Demo of IEEEtran.cls for IEEE Journals}

\maketitle

\makeatletter

\begin{abstract}

3D point cloud registration is fragile to outliers, which are labeled as the points without corresponding points. To handle this problem, a widely adopted strategy is to estimate the relative pose based only on some accurate correspondences, which is achieved by building correspondences on the identified inliers or by selecting reliable ones. However, these approaches are usually complicated and time-consuming. By contrast, the virtual point-based methods learn the \underline{v}irtual \underline{c}orresponding \underline{p}oints (VCPs) for all \textit{source} points uniformly without distinguishing the outliers and the inliers. Although this strategy is time-efficient, the learned VCPs usually exhibit serious collapse degeneration due to insufficient supervision and the inherent distribution limitation. In this paper, we propose to exploit the best of both worlds and present a novel robust 3D point cloud registration framework. We follow the idea of the virtual point-based methods but learn a new type of virtual points called \underline{r}ectified virtual \underline{c}orresponding \underline{p}oints (RCPs), which are defined as the point set with the same shape as the \textit{source} and with the same pose as the \textit{target}. Hence, a pair of consistent point clouds, \ie \textit{source} and RCPs, is formed by rectifying \underline{V}CPs to \underline{R}CPs (VRNet), through which reliable correspondences between \textit{source} and RCPs can be accurately obtained. Since the relative pose between \textit{source} and RCPs is the same as the relative pose between \textit{source} and \textit{target}, the input point clouds can be registered naturally. Specifically, we first construct the initial VCPs by using an estimated soft matching matrix to perform a weighted average on the \textit{target} points. Then, we design a correction-walk module to learn an offset to rectify VCPs to RCPs, which effectively breaks the distribution limitation of VCPs. Finally, we develop a hybrid loss function to enforce the shape and geometry structure consistency of the learned RCPs and the \textit{source} to provide sufficient supervision. Extensive experiments on several benchmark datasets demonstrate that our method achieves advanced registration performance and time-efficiency simultaneously. 
\end{abstract}

\begin{IEEEkeywords}
Point cloud registration, distribution degeneration, rectified virtual corresponding points, correction-walk module, hybrid loss function.
\end{IEEEkeywords}

\IEEEpeerreviewmaketitle

%%%%%%%%%%%%%%%%%%%% main text %%%%%%%%%%%%%%%%%%%%%
\section{Introduction} \label{1-introduction}
%%%%%%%%%%%%%%%%%%%%%%%%%%%%%%%%%%%%%%%%%%%%%%%%%%%%

As an important data type to describe 3D scene, point cloud has received considerable attention \cite{charles_pointnet_cvpr_2017,wang_dgcnn_tog_2019,li_tcsvt_multi_2021}. More importantly, 3D point cloud registration, as a key problem in 3D computer vision community, has been adopted in various applications, such as 3D reconstruction \cite{deschaud_imls_icra_2018,zhang_loam_rss_2014}, autonomous driving \cite{yang_robust_iros_2018,wan_robustlocalization_icra_2018}, simultaneous localization and mapping (SLAM) \cite{ding_deepmapping_cvpr_2019}, locating 3d object \cite{pahwa_locate_tcsvt_2018}, point cloud code \cite{Mekuria_codec_tcsvt_2017}.
The point cloud registration task aims at solving the relative pose of 6 degrees of freedom to optimally align the two input point clouds, \ie \textit{source} and \textit{target}, which has been studied for many years. 
Many traditional approaches have achieved remarkable performance. For example, \cite{besl_icp_pami_1992,huang_ctf_tcsvt_2018} advocate using a coarse-to-fine strategy to solve for accurate 3D registration.
Recently, benefiting from the rise of deep learning technique, learning-based 3D point cloud registration has become a new hot spot, where correspondences-free methods (\eg \cite{aoki_ptlk_cvpr_2019,huang_featuremetric_cvpr_2020}) and correspondences-based methods (\eg \cite{lu_deepvcp_iccv_2019,3dlocal_Deng_CVPR_19}) are developed depending on whether the correspondences are explicitly built or not.

\begin{figure}[t]
    \centering
    \includegraphics[width=0.99\linewidth]{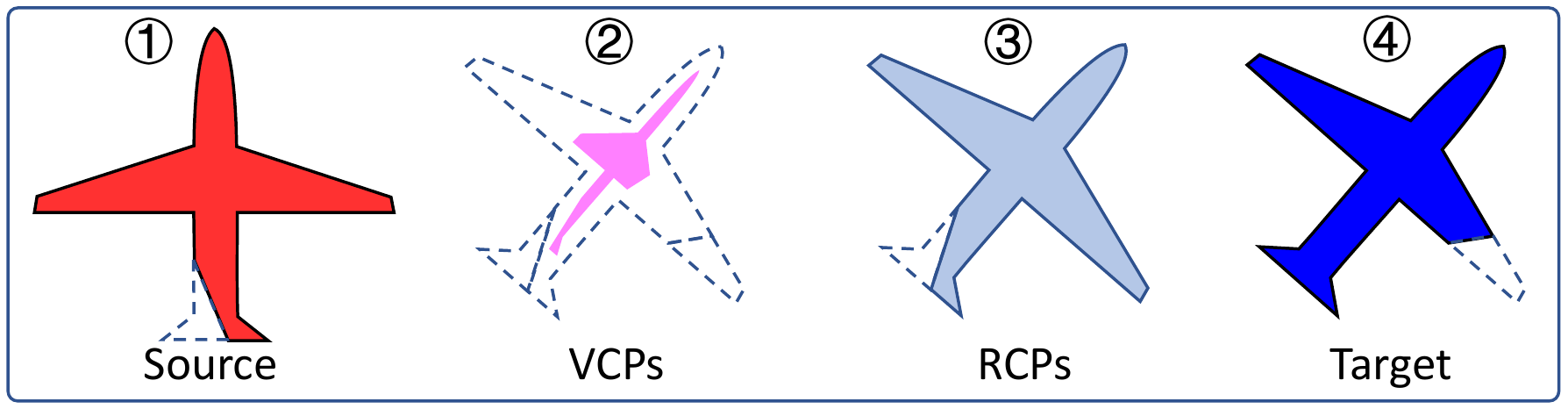}\vspace{-1.5mm}
    \caption{Illustration of our VRNet. \ding{172} \textit{source} and \ding{175} \textit{target} have different poses and different shapes (broken tail and wing in \textit{source} and \textit{target} respectively). The existing methods will learn degenerated VCPs indicated by the pink in \ding{173} (\cf \figref{Fig:comparison}). Conversely, our VRNet devotes to learning the RCPs indicated by \ding{174}, which maintain the same shape as the \textit{source} and the same pose as the \textit{target}, by unfolding VCPs and rectifying the partiality of the wing. Hence, the reliable correspondences of these consistent point clouds, \ie \textit{source} and RCPs, can be obtained easily since the influence of outliers has been eliminated. Further, the relative pose between \textit{source} and RCPs can be solved accurately, which is same as the relative pose between \textit{source} and \textit{target}.}
    \label{Fig:rcps}
    \vspace{-5mm}
\end{figure}

\begin{figure}[!t]
    \centering
    \includegraphics[width=0.65\linewidth]{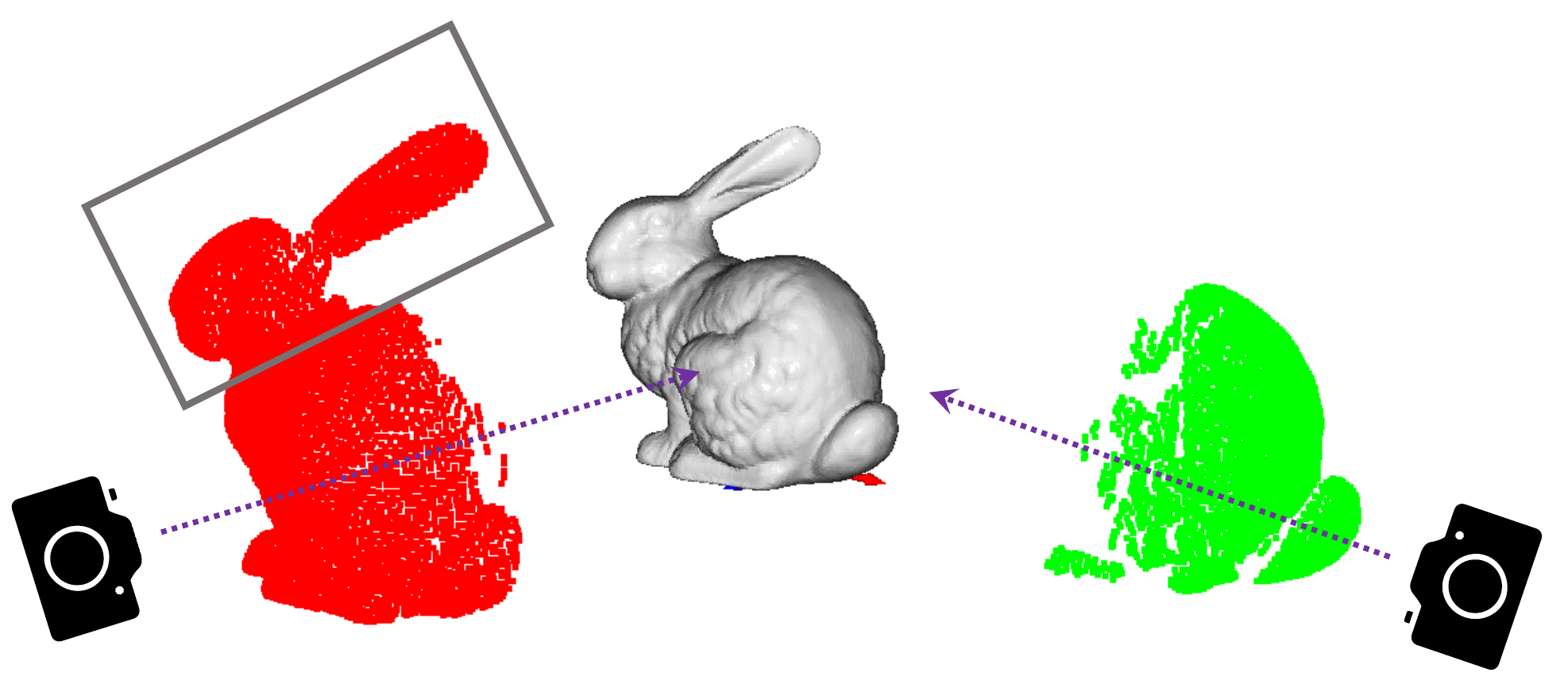}\vspace{-1.5mm}
    \caption{Illustration of the distribution limitation of VCPs. The red and green represent the \textit{source} and the \textit{target} respectively. In this case, only a part of corresponding points can be fitted by the VCPs, which are generated by performing the weighted average on the \textit{target}. And the corresponding points of the \textit{source} points marked by the box can never be fitted since the distribution of the VCPs is limited in the convex set of the \textit{target}.}
    \label{Fig:refinement}
    \vspace{-5mm}
\end{figure}

However, the widespread presence of outliers, \ie the points without corresponding points in the paired point clouds, has always been a significant challenge for both correspondences-free and correspondences-based point cloud registration methods.
Note that the essence of the correspondences-free methods is to estimate the relative pose by comparing the global representations of two input point clouds \cite{aoki_ptlk_cvpr_2019,sarode_pcrnet_arxiv_2019,huang_featuremetric_cvpr_2020}. Thus, the outliers are destructive for these correspondences-free methods because the difference between their global representations can no longer indicate their pose difference (\ie the relative pose) accurately.
In other words, the shape differences due to the outliers, \eg, the head of rabbit only exists in the \textit{source} without corresponding points as illustrated in \figref{Fig:refinement}, also contribute to the difference in their global representations.
As a result, the correspondences-based methods are gaining more and more attention, which advocate going further to deal with the disturbance of the outliers by building some accurate correspondences from the contaminated input point clouds.

\begin{figure}[!t]
    \centering
    \includegraphics[width=0.99\linewidth]{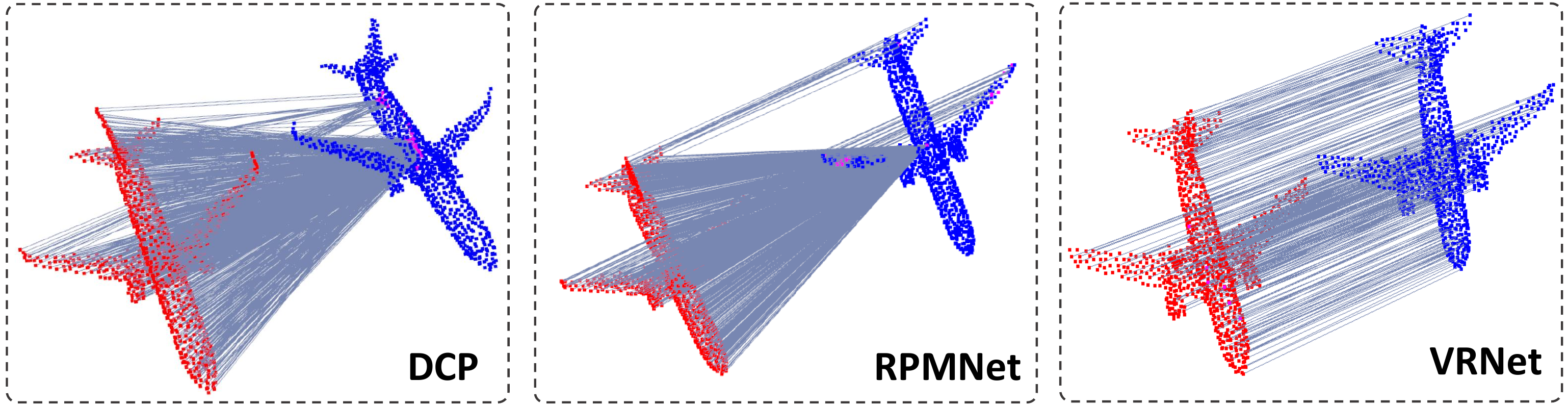}
    \caption{The degeneration of the learned corresponding points. The results are generated from the consistent input point clouds for a clear comparison of loss function, where the effect of the distribution limitation is excluded naturally. Red and blue represent the \textit{source} and the \textit{target} respectively. Pink indicates the learned corresponding points. The matching lines connect the \textit{source} points and the corresponding points. 
    Due to insufficient supervision, the learned corresponding points of DCP \cite{wang_dcp_iccv_2019} and RPMNet \cite{yew_rpmnet_cvpr_2020} degenerate seriously. Our VRNet achieves much better performance in which the learned corresponding points maintain the original shape and geometry structure due to the proposed hybrid loss function.}
    \label{Fig:comparison}
    \vspace{-\baselineskip}
\end{figure}

To this end, a virtual point-based strategy is employed \cite{wang_dcp_iccv_2019,lu_deepvcp_iccv_2019,yew_rpmnet_cvpr_2020}. It advocates the use of \underline{v}irtual \underline{c}orresponding \underline{p}oints (VCPs), which are constructed by performing weighted average on the \textit{target}, instead of the real points in the \textit{target}. 
% instead of the real points for all \textit{source} points without distinguishing inliers and outliers. 
However, as shown in \figref{Fig:comparison}, the correspondences brought by this strategy are not reliable because the learned VCPs exhibit serious collapse degeneration and lose the shape and geometry structure, which has been proved in \cite{hpnet_arxiv_2021}. 
Two reasons exist for these degenerations: 1) the existing supervision usually focuses on the relative pose only, which is insufficient and more than one feasible solutions exist; 2) the distribution of the virtual points is limited in the \textit{target} due to the weighted average operation as depicted in \figref{Fig:refinement}. 
Nevertheless, it is worth mentioning that because of the uniform treatment of the \textit{source} points without the complicated distinguishing process, the virtual point-based approaches usually own a high time-efficiency.
Meanwhile, real point-based approaches have received more and more attention recently, which build reliable correspondences on real points. To this end, a natural idea is to identify the inliers and then build correspondences on these inliers only. 
PRNet \cite{wang_prnet_nips_2019} proposes to select the points with more obvious feature as the inliers, however, this operation is neither interpretable nor persuasive.
\cite{predator_Huang_2021_CVPR} utilizes the attention mechanism to recognize inliers, but this is operationally complex as well as time-inefficient.
Real point-based approaches also devote to selecting reliable correspondences from the constructed initial correspondences. In \cite{pais_3dregnet_cvpr_2020,choy_dgr_cvpr_2020,probst_consensusMax_cvpr_2019}, the correspondences are selected based on the learned reliability weight of each correspondence. RANSAC is also widely adopted to select consistent correspondences \cite{3dlocal_Deng_CVPR_19,choy_dgr_cvpr_2020}. However, these real point-based methods often struggle with high computational efficiency and the ability to obtain reliable correspondences. 

Due to the respective limitations of both virtual point-based and real point-based approaches, we point out that constructing the reliable corresponding points of all the \textit{source} points uniformly without distinguishing the inliers and the outliers can effectively incorporate their advantages. By this way, high time-efficiency and high accuracy can be achieved at the same time. For this goal, we propose to learn a new type of virtual points called \underline{r}ectified virtual \underline{c}orresponding \underline{p}oints (RCPs), which are defined as the point set with the same shape as the \textit{source} and with the same pose as the \textit{target}, as shown in \figref{Fig:rcps}. 
Therefore, a pair of consistent point clouds, \ie \textit{source} and RCPs, can be formed to eliminate the influence of outliers via rectifying \underline{V}CPs to \underline{R}CPs (VRNet). Then one can easily yield reliable correspondences to solve for the relative pose between the \textit{source} and RCPs, \ie the relative pose between the \textit{source} and \textit{target}.
Our VRNet consists of two main steps. Firstly, we construct the initial VCPs by using a soft matching matrix to perform the weighted average on the \textit{target} point cloud. Secondly, we propose a correction-walk module to learn an offset to rectify VCPs to RCPs, which breaks the inherent distribution limitation of original VCPs. Besides, a novel hybrid loss function is proposed to enhance the consistency of shape and geometric structure between the learned RCPs and the \textit{source} point cloud. 
The proposed hybrid loss function consists of \textit{corresponding point supervision}, \textit{local motion consensus}, \textit{geometry structure supervision}, and \textit{amendment offset supervision}. It is proved to be effective to supervise the entire network from the perspectives of the inliers distribution, the consistency of local and global motions, the geometry structure, \etc. 
Finally, we evaluate the proposed VRNet through extensive experiments on synthetic and real data, achieving state-of-the-art registration performance.
Furthermore, our method is time-efficient since it circumvents the complicated processes of inliers determination and correspondences selection.

Our contributions can be summarized as follows:
\begin{itemize}
\setlength{\itemsep}{0pt}
\setlength{\parsep}{0pt}
\setlength{\parskip}{0pt}
    \item[1).] We propose a point cloud registration method named VRNet to guarantee high accuracy and high time-efficiency. We present a new type of virtual points called RCPs, which maintain the same shape as the \textit{source} and the same pose as the \textit{target}, to help build reliable correspondences. 
	\item[2).] We design a novel correction-walk module in our VRNet to learn an offset to break the distribution limitation of the initial VCPs. Besides, a hybrid loss function is proposed to enhance the rigidity and geometric structure consistency between the learned RCPs and the \textit{source}.
	\item[3).] Remarkable results on benchmark datasets validate the superiority and effectiveness of our proposed method for robust 3D point cloud registration.
\end{itemize}

%%%%%%%%%%%%%%%%%%%%%%%%%%%%%%%%%%%%%%%%%%%%%%%
\section{Related work} \label{sec:relatedwork}
%%%%%%%%%%%%%%%%%%%%%%%%%%%%%%%%%%%%%%%%%%%%%%%

Employing the deep learning technique to the 3D point cloud registration task has received widespread attention recently. In this section, we provide a brief review of the learning-based point cloud registration methods. And the detailed summary of traditional point cloud registration methods has been provided in \cite{ruslinkiewicz_efficientVariantsIcp_3DDIM_2001,Fran_prc_2015}.

\subsection{Correspondences-free methods}
Deep learning technique provides a new perspective for the 3D point cloud registration task, \ie solving the rigid transformation by comparing the holistic representations of the \textit{source} point cloud and the \textit{target} point cloud. This kind of method is usually called the correspondences-free method and consists of two main steps: global feature extraction and rigid motion solving. 
PointNetLK \cite{aoki_ptlk_cvpr_2019} represents a pioneer, which uses PointNet \cite{charles_pointnet_cvpr_2017} to extract the global features of the \textit{source} and \textit{target}. And then a modified LK algorithm is designed to solve the rigid transformation from the difference between these two global features. 
A similar work, PCRNet \cite{sarode_pcrnet_arxiv_2019}, proposes to replace the modified LK algorithm with a regression strategy, which brings more accurate registration results. Huang \etal \cite{huang_featuremetric_cvpr_2020} propose a more effective global feature extractor inspired by the reconstruction methods \cite{zhao_capsule_cvpr_2019,yang_foldingnet_cvpr_2018}, in which an encoder-decoder network is designed to learn a more comprehensive global representation.
However, when outliers exist, there are significant differences in shape and geometry structure between the \textit{source} and \textit{target} in addition to the poses, thus the correspondences-free method usually fails to obtain accurate registration results.

\subsection{Correspondences-based methods}

Correspondences-based methods are built upon the correspondences, which consist of two main steps: correspondences building and rigid transformation estimation. 
Comparing with the point-to-plane correspondences, plane-to-plane correspondences, \etc, the point-to-point correspondences are the most common in correspondences-based methods. Among them, feature extractor and reliable correspondences building modules are depthly explored for more accurate registration.

\noindent\textbf{Feature extractors.}
A number of effective point feature learning methods are employed in the point cloud registration task to obtain more accurate and suitable descriptors for more accurate alignment.
3DFeat-Net \cite{yew_3dfeatnet_eccv_2018} utilizes a set abstraction module proposed by \cite{charles_pointnet2_nips_2017} to summarize the local geometric structure. 
DCP \cite{wang_dcp_iccv_2019} uses DGCNN \cite{wang_dgcnn_tog_2019} and Transformer \cite{vaswani_attention_nips_2017} to learn the task-specific features. 
Different from the above methods, RPMNet \cite{yew_rpmnet_cvpr_2020} proposes to use a 10D hybrid feature representation, where the normal is additionally used besides the 3D coordinate.
To handle the large-scale scene data, DGR \cite{choy_dgr_cvpr_2020} uses a fully connected network \cite{choy_fcgf_iccv_2019} to extract features.  
Deng \etal propose to learn a globally informed 3D local feature in \cite{PPFNet_Deng_2018_CVPR}. 
Via an unsupervised learning network, PPF-FoldingNet \cite{PPFFolding_Deng_2018_ECCV}, a rotation invariant 3D local descriptor is designed. 
In \cite{perfect_Gojcic_cvpr19}, a rotation invariant feature, voxelized smoothed density value, is used for point matching. D3Feat \cite{D3Feat_Bai_2020_CVPR} leverages KPConv \cite{thomas_kpconv_iccv_19} to predict both a detection score and a feature for each 3D point, which helps to detect key points and extract effective features at the same time. With the advent of deep learning techniques, learning-based feature extractors have become standard modules that are easy to integrate.

\begin{figure*}[!t]
	\centerline{\includegraphics[width=0.9\linewidth]{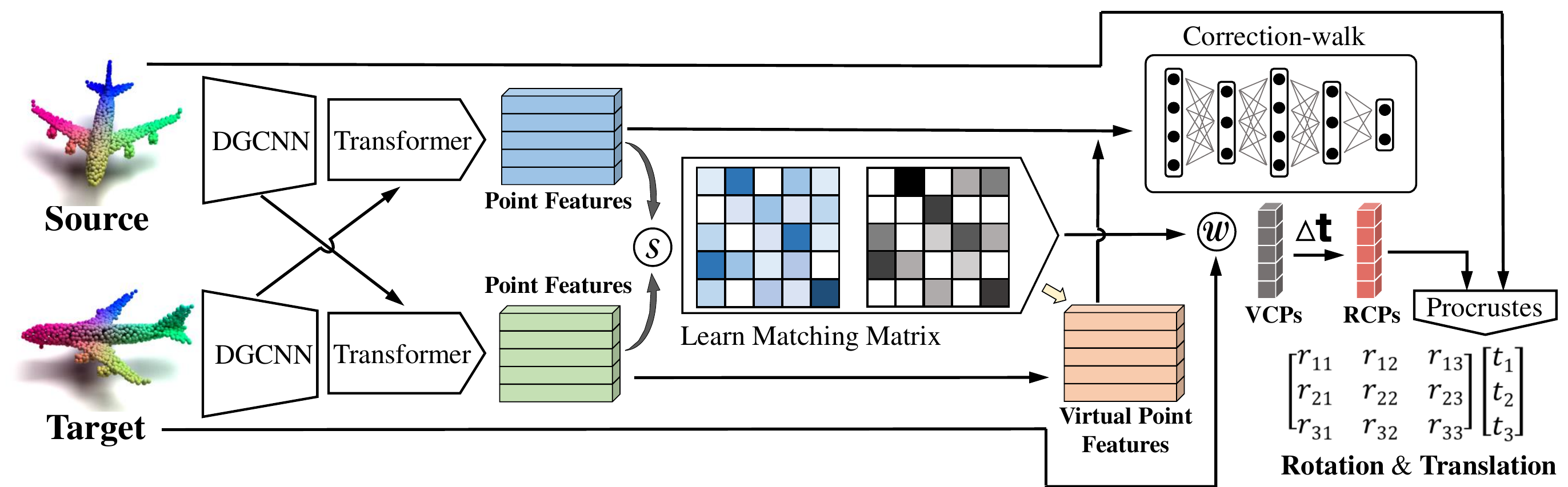}}
	\caption{The network architecture of our proposed VRNet. Given the \textit{source} and \textit{target}, DGCNN and Transformer are applied to extract point features. Then, a soft matching matrix is achieved based on the constructed similarity matrix. Virtual corresponding points and corresponding point features are obtained by using the matching matrix to perform the weighted average on the \textit{target} point cloud and the \textit{target} point features respectively. To break the distribution limitation, a correction-walk module is proposed to learn the offset to amend the VCPs to the desired RCPs. Finally, the rigid transformation is solved by the Procrustes algorithm. The network is supervised by the proposed hybrid loss function, which enforces the rigidity and geometry structure consistency between the learned RCPs and the \textit{source} point cloud.}
	\label{Fig:network}
\end{figure*}

\noindent\textbf{Correspondences building.} 
Note that the accuracy of the correspondences is more important than the number of correspondences in the registration task, constructing some reliable correspondences for robust registration has become a widely accepted strategy. 
To this end, some methods propose to distinguish inliers and outliers, and then solely build correspondences on the identified inliers. 
For example, Predator \cite{predator_Huang_2021_CVPR} proposes an overlap attention module to recognize the inliers. PRNet \cite{wang_prnet_nips_2019} selects the points with more obvious features as inliers.
Besides, selecting reliable correspondences from the constructed initial correspondences is also an effective method.
3DRegNet \cite{pais_3dregnet_cvpr_2020}, DGR \cite{choy_dgr_cvpr_2020}, and consensus maximization method \cite{probst_consensusMax_cvpr_2019} concentrate on estimating the reliability weight of each correspondence. 
The 3DRegNet derives each pair of points individually, and DGR considers the neighbor information by high dimension convolution. 
The consensus maximization method is unsupervised using the principle of maximizing the number of consistent correspondences. Moreover, consistent correspondences are also selected based on the RANSAC \cite{3dlocal_Deng_CVPR_19}.
% The consensus maximization method is unsupervised utilizing the principle of the number of inliers maximization. 
However, these operations are complicated and time-consuming. 
Thus, an alternative virtual point-based strategy is proposed, which constructs the correspondences for all \textit{source} points without distinguishing inliers and outliers using virtual points. In DeepVCP \cite{lu_deepvcp_iccv_2019}, these virtual corresponding points are constructed by the weighted average on the points generated based on the prior transformation. In contrast, all the points in the \textit{target} point cloud are used in DCP \cite{wang_dcp_iccv_2019}. However, this virtual point-based outlier processing strategy suffers from the serious degeneration due to two reasons, \ie the insufficient supervision and the distribution limitation of the learned virtual corresponding points.

\noindent\textbf{Rigid transformation estimation}. The Procrustes algorithm \cite{gower_procrustes_1975} is the most common strategy to solve the rigid transformation \cite{wang_dcp_iccv_2019, wang_prnet_nips_2019,lu_deepvcp_iccv_2019,yew_rpmnet_cvpr_2020}, which has been proved to be optimal based on the correct correspondences. In addition, the direct regression of motion parameters has also received widespread attention in recent years \cite{pais_3dregnet_cvpr_2020,sarode_pcrnet_arxiv_2019,3dlocal_Deng_CVPR_19}.

%%%%%%%%%%%%%%%%%%%%%%%%%%%%%%%%%%%%%%%%%%%%%%%%%%%%
\section{Proposed Method} \label{4-method}
%%%%%%%%%%%%%%%%%%%%%%%%%%%%%%%%%%%%%%%%%%%%%%%%%%%%

\subsection{Preliminaries}
3D point cloud registration devotes to estimating the rigid transformation best aligning the two given point clouds $\mathbf{X} = [{\mathbf{x}_i}] \in \mathbb{R}^{3\times N_\mathbf{X}}$, and $\mathbf{Y} = [\mathbf{y}_j] \in \mathbb{R}^{3\times N_\mathbf{Y}}$, where ${\mathbf{x}_i} \in \mathbb{R}^3$, $\mathbf{y}_j \in \mathbb{R}^3$, $N_\mathbf{X}$ and $N_\mathbf{Y}$ are the numbers of points in $\mathbf{X}$ and $\mathbf{Y}$ respectively and $N_\mathbf{X}$, $N_\mathbf{Y}$ do not need to be equal. Usually, $\mathbf{X}$ and $\mathbf{Y}$ are called the \textit{source} point clouds and the \textit{target} point clouds, respectively. In this paper, we model the rigid transformation by the rotation matrix $\mathbf{R} \in SO(3)$ and the translation vector $\mathbf{t} \in \mathbb{R}^3$.

Furthermore, point matching is a key problem in the 3D point cloud registration task, which is usually tackled by solving a binary matching matrix $\mathbf{M}=[m_{ij}]_{{N}_\mathbf{X} \times {N}_\mathbf{Y}}$, where $m_{ij} \in \{0,1\}$, \ie 
\begin{equation}
    m_{ij} = 
        \begin{cases}
        1 & \text{if $\boldsymbol{x}_i$ and $\boldsymbol{y}_j$ are matched,} \\
        0 & \text{otherwise.}
        \end{cases}
\label{Eq:matching_matrix}
\end{equation} 
Within the virtual point-based methods, the matching matrix $\mathbf{M}$ is relaxed to $[0,1]^{{N}_\mathbf{X} \times {N}_\mathbf{Y}}$, where $m_{ij}$ represents the matching probability between the point $\mathbf{x}_i$ and the point $\mathbf{y}_j$, and $\mathbf{M}$ is called the soft matching matrix.

\subsection{VRNet architecture} \label{3-1-pipeline}
%%%%%%%%%%%%%%%%%%%%%%%%%%%%%%%%%%%%%%%%%%%%%%%%%%%%
We advocate constructing the reliable corresponding points of all the \textit{source} points uniformly without distinguishing the inliers and the outliers to ensure both high time-efficiency and high accuracy. 
To this end, we propose to learn a new type of virtual points called RCPs, which are defined as the point set with the same shape as the \textit{source} and with the same pose as the \textit{target}. By this way, a pair of consistent point clouds, \ie \textit{source} and RCPs, is produced to eliminate the influence of outliers. Meanwhile, rectifying \underline{V}CPs to \underline{R}CPs facilitates generating reliable correspondences to solve for the relative pose between the \textit{source} and RCPs, which is same with the relative pose between the \textit{source} and \textit{target}.
The entire architecture of our VRNet is illustrated in \figref{Fig:network}. First, the initial VCPs are constructed. Then, the RCPs are achieved by learning a rectified offset in the correction-walk module. Finally, the rigid transformation is estimated by the Procrustes algorithm. We introduce these procedures in detail as follows.

\noindent\textbf{VCPs construction.} 
Inspired by DCP \cite{wang_dcp_iccv_2019}, we construct the virtual corresponding points by using the matching matrix $\mathbf{M}$ to perform weighted average on the \textit{target} point cloud $\mathbf{Y}$. To this end, we apply the ``DGCNN + Transformer'' as our feature extractor at first. Specifically, a shared DGCNN \cite{wang_dgcnn_tog_2019} is employed to compute the initial point features for the two input point clouds because it can achieve informative representation by summarizing the neighbor information through edge convolution operation. Besides, inspired by the recent success of attention mechanism, the Transformer module \cite{vaswani_attention_nips_2017} is also used to learn co-contextual information of the \textit{source} point clouds and the \textit{target} point clouds. Formally, the DGCNN feature extraction can be summarized as,
\begin{equation}
    \mathbf{F}_{\mathbf{x}_i}^\ell = \text{maxpool}(\textbf{MLP}_\alpha(\text{cat}(\mathbf{F}_{\mathbf{x}_i}^{\ell-1}, \mathbf{F}_{\mathbf{x}_{ik}}^{\ell-1}))),\ \mathbf{F}_{\mathbf{x}_{ik}}\in \mathbb{N}({\mathbf{F}_{\mathbf{x}_i}}),
\label{equ:dgcnn}
\end{equation}
where $\ell$ represents the $\ell\text{-th}$ layer of edge convolution. $\mathbb{N}({\mathbf{F}_{\mathbf{x}_i}})$ denotes the K-nearest neighbors of ${\mathbf{F}_{\mathbf{x}_i}}$ in thhe feature space with the pre-defined parameter $K$, \ie $k\in [1,K]$.  The initial point feature is the original 3D coordinate. $\textbf{MLP}_\alpha$ is a multi-layer perceptron (MLP) network parameterized by $\alpha$. $\text{cat}(\cdot,\cdot)$ represents the concatenation operation and $\text{maxpool}(\cdot)$ represents the max-pooling operation.
After several edge convolutions, point features are achieved and denoted as $\mathbf{F}_\mathbf{X}=[\mathbf{F}_{\mathbf{x}_i}]\in \mathbb{R}^{N_\mathbf{X}\times c}$, $\mathbf{F}_\mathbf{Y}=[\mathbf{F}_{\mathbf{y}_j}]\in \mathbb{R}^{N_\mathbf{Y}\times c}$ where $c$ is the pre-defined feature dimension. Then, the Transformer module is applied as,
\begin{equation}
\left\{ {\begin{aligned}
	\Phi_\mathbf{X} &= \mathbf{F}_\mathbf{X} + \eta_1(\mathbf{F}_\mathbf{X}, \mathbf{F}_\mathbf{Y})\\
	\Phi_\mathbf{Y} &= \mathbf{F}_\mathbf{Y} + \eta_2(\mathbf{F}_\mathbf{Y}, \mathbf{F}_\mathbf{X})
	\end{aligned}} \right. ,
\end{equation}
where $\eta_1: \mathbb{R}^{N_\mathbf{X}\times c} \times \mathbb{R}^{N_\mathbf{Y}\times c} \to \mathbb{R}^{N_\mathbf{X} \times c}$ and $\eta_2: \mathbb{R}^{N_\mathbf{Y}\times c} \times \mathbb{R}^{N_\mathbf{X}\times c} \to \mathbb{R}^{N_\mathbf{Y} \times c}$ represent the Transformer function. The features of $\mathbf{x}_i$ and $\mathbf{y}_j$ are denoted as $\Phi_{\mathbf{x}_i}$ and $\Phi_{\mathbf{y}_j}$, so $\Phi_\mathbf{X}\in \mathbb{R}^{N_\mathbf{X}\times c}$ and $\Phi_\mathbf{Y}\in \mathbb{R}^{N_\mathbf{Y}\times c}$ indicate the final point features of all \textit{source} points and \textit{target} points.

Then, we take the scaled dot product attention metric to calculate the similarity matrix $\mathbf{S}=[s_{ij}]_{N_{\mathbf{X}} \times N_{\mathbf{Y}}}$, where
\begin{equation}
    s_{ij} = \Phi_{\mathbf{x}_i} \Phi_{\mathbf{y}_j}^\text{T} / \sqrt{c}.
\end{equation}
Next, row-wise $\text{softmax}$ operation is employed for final soft matching matrix $\mathbf{M} = \text{softmax}(\mathbf{S})$. Then, VCPs of the \textit{source} point cloud are achieved as $\mathbf{Y}^{\prime} = \mathbf{Y}\mathbf{M}^\mathrm{T}$, $\mathbf{Y}^{\prime} \in \mathbb{R}^{3\times N_\mathbf{X}}$.

\noindent\textbf{RCPs construction by correction-walk.} 
%%%%%%%%%%%%%%%%%%%%%%%%%%%%%%%%%%%%%%%%%%%%%%%%%%%%%%%%%%%%
Constructing the VCPs to fit the real corresponding points is the fundamental principle of existing virtual point-based methods. This idea makes sense when the real corresponding points are surrounded or overlapped by the \textit{target} points. In this case, these points can be fitted by learning a soft matching matrix $\mathbf{M}\in [0,1]^{N_\mathbf{X} \times N_\mathbf{Y}}$.
However, note that the weighted average on \textit{target} using the soft matching matrix $\mathbf{M}$ can only cover a convex set in 3D space, some real corresponding points of outliers cannot be fitted since they are outside the range of this convex set. A typical example is presented in \figref{Fig:refinement}. This shortage is common in practice and results in many wrong correspondences. To break this distribution limitation of the VCPs, we propose a correction module called correction-walk to learn the offsets to rectify VCPs to RCPs.

For VCPs $\mathbf{Y}^{\prime}$, we construct the corresponding virtual point features analogously, \ie $\Phi_{\mathbf{Y}^{\prime}}= \Phi_\mathbf{Y} \mathbf{M}^\mathrm{T}$. 
Then, we formulate $\mathbf{E}=\text{cat}(\Phi_{\mathbf{X}},\Phi_{\mathbf{Y}^{\prime}}) \in \mathbb{R}^{N_\mathbf{X}\times 2c }$, where $\text{cat}(\cdot,\cdot)$ denotes the concatenation operation. Because we advocate rectifying the degenerated VCPs to the RCPs, whose shape is defined to be the same as the \textit{source}. Thus, according to the feature differences between the VCPs and the \textit{source}, \ie $\Phi_{\mathbf{X}}$ and $\Phi_{\mathbf{Y}^{\prime}}$, it is expected that the offsets from VCPs to RCPs are generated from $\mathbf{E}$. Therefore, $\mathbf{E}$ is called seeds in our paper.
The proposed correction-walk module learns the correction displacement from the seeds. Specifically, the correction-walk module is implemented by another MLP network, which consumes the seeds and outputs the Euclidean space offset $\Delta \mathbf{t}_{\mathbf{X}} \in \mathbb{R}^{3\times N_\mathbf{X}}$, \ie
\begin{equation}
    \Delta \mathbf{t}_{\mathbf{X}} = \mathbf{MLP}_\beta(\mathbf{E}).
\end{equation}
Thus, the final learned RCPs are produced by adding the learned offset to VCPs, \ie $\mathbf{Y}^{\prime\prime} = \mathbf{Y}^{\prime} + \Delta \mathbf{t}_{\mathbf{X}}$, $\mathbf{Y}^{\prime\prime} \in \mathbb{R}^{3\times N_\mathbf{X}}$.

\noindent\textbf{Rigid transformation estimation.}
After matching the points in $\mathbf{X}$ with the points in $\mathbf{Y}^{\prime\prime}$, the rigid transformation between the \textit{source} and RCPs can be solved in closed-form by the Procrustes algorithm \cite{gower_procrustes_1975}. Because the pose of RCPs is same as the \textit{target}, the desired final relative pose is obtained naturally. Specifically, $\mathbf{H} = \sum_{i = 1}^{N_\mathbf{X}} (\mathbf{x}_{i} -\mathbf{\bar{x}} ) (\mathbf{y}_{i}^{\prime\prime} - \mathbf{\bar{y^{\prime\prime}}})^\mathrm{T}$, where $\mathbf{ \bar{x}}$ and $\mathbf{\bar{ y^{\prime\prime}}}$ are the centers of $\mathbf{X}$ and $\mathbf{{Y}^{\prime\prime}}$. Then, by using the singular value decomposition (SVD) to decompose $\mathbf{H} = \mathbf{UDV}^\mathrm{T}$, we obtain the final rigid transformation as,
\begin{equation}
    \left\{
    \begin{aligned}
    \mathbf{R} &= \mathbf{V}\mathbf{U}^\mathrm{T} \\
    \mathbf{t} &= \mathbf{-R} \mathbf{\bar x} + \mathbf{\bar{y^{\prime\prime}}}
    \end{aligned}
    \right. .
\end{equation}

Note that the Procrustes algorithm makes sense based on the correct correspondences. Hence, this rigid transformation estimation solvers would be problematic or even invalid if the learned virtual corresponding points degenerate. Unfortunately, this trap has been generally ignored. In this paper, we propose the RCPs, which rectify this inherent drawback of original VCPs to guarantee the reliability of constructed correspondences for the final relative pose estimation. 

%%----------------------------------------------------------
\subsection{Loss function} \label{3-2-loss}
%%%%%%%%%%%%%%%%%%%%%%%%%%%%%%%%%%%%%%%%%%%%%%%%%%%%%%%%%%%%
In existing virtual point-based point cloud registration methods, \eg DCP \cite{wang_dcp_iccv_2019}, the loss function usually only focuses on supervising the final rigid transformation. Due to this insufficient constraint, the distribution of the learned corresponding points degenerates as shown in \figref{Fig:comparison}. 
To solve this problem, we propose a novel hybrid loss function, which devotes to driving the learned RCPs and the \textit{source} point cloud keep consistent in terms of the rigidity and geometry structure. 

\noindent\textbf{Corresponding point supervision.}
This supervised loss function concentrates on the predicted matching matrix $\mathbf{M}$. Although $\mathbf{M}$ is a soft probability matrix, it is enforced to the ground truth binary matching matrix to keep rigidity. Herein, we design the loss function as:
\begin{equation}
\mathcal{L}_0 =  - \frac{{\sum_{i=1}^{N_X} \sum_{j=1}^{N_Y} \left( {{m}_{ij}^\text{pred} {m}_{ij}^\text{gt}} \right)}}{{\sum_{i=1}^{N_X} \sum_{j=1}^{N_Y} \left( {{m}_{ij}^\text{gt}} \right)}},
\label{Eq:single_loss}
\end{equation}
where the superscript ``$\text{pred}$'' and ``$\text{gt}$'' represent the prediction and ground truth respectively, $m_{ij}$ is the entry of the matching matrix $\mathbf{M}$. However, if $\mathbf{x}_i$ is outlier, $m_{ij}^\text{gt}=0$ for all $j=1,...,N_{\mathbf{Y}}$. At present, $m_{ij}^\text{pred}$ is divergent, \ie\ $\mathcal{L}_0$ can only supervise the inliers but ignore the outliers. \textit{By corresponding point supervision, we emphasize the distribution of inliers. }

\noindent\textbf{Local motion consensus.}
To guarantee the predicted corresponding points to keep rigidity, the rigid motion estimated according to all correspondences and the rigid motion estimated according to the correspondences subset should be the same, that is \textit{local motion consensus}. Specifically, because the \textit{source} point cloud $\mathbf{X}$ and the predicted RCPs $\mathbf{Y}^{\prime\prime}$ are matched, the correspondences set can be obtained as $\Omega = \{(\mathbf{x}_i,\mathbf{y}_i^{\prime\prime})|i \in [1,N_\mathbf{X}], \mathbf{x}_i \in \mathbf{X}, y_i^{\prime\prime} \in \mathbf{Y}^{\prime\prime}\}$. The global optimal rigid motion can be solved by the Procrustes algorithm \cite{gower_procrustes_1975} based on $\Omega$, notated as $\mathbf{R}$, $\mathbf{t}$. Then, we select $G$ subsets of correspondences randomly, \ie, $\Omega_g \subset \Omega$, $g\in [1,G]$. The size of each subset is $|\Omega_g| \geq 3$. At present, the local rigid motion $\mathbf{R}_g,\mathbf{t}_g$ can also be solved according to $\Omega_g$ by the Procrustes. Ideally, $\mathbf{R}_g,\mathbf{t}_g$ should be the same as $\mathbf{R},\mathbf{t}$. Based on this observation, we define an unsupervised loss function as:
\begin{equation}
\mathcal{L}_1 =  \frac{1}{G}\sum_{g=1}^{G} (\text{rmse}(\textbf{R}_g^\textrm{T}\textbf{R}, \mathbf{I}_3)+\text{rmse}(\textbf{t}_g,\textbf{t})),
\label{Eq:local_loss}
\end{equation}
where $\mathbf{I}_3$ is a 3-order identity matrix, $\text{rmse}(\cdot,\cdot)$ is the root mean squared error. $\mathcal{L}_1$ should converge to 0 ideally. \textit{By the local motion consensus, we drive the motion of each local part to be consistent with the global motion}.

\noindent\textbf{Geometry structure supervision.}
In this part, the \textit{source} point cloud $\mathbf{X}$ and the predicted RCPs $\mathbf{Y}^{\prime\prime}$ are formulated as two graphs, respectively. Each point is a node and the distance of arbitrary two points is an edge. Obviously, the edge of $\mathbf{x}_i, \mathbf{x}_j \in \mathbf{X}$ should be the same as the edge of the corresponding two points $\mathbf{y}_i^{\prime\prime}, \mathbf{y}_j^{\prime\prime} \in \mathbf{Y}^{\prime\prime}$. 
Here, we denote the edge matrix of the \textit{source} as $\mathbf{D}$,
\begin{equation}
    \mathbf{D} = \left[ {\begin{array}{*{20}{c}}
0&{d_{1,2}}&{\cdots}&{d_{1,N_\mathbf{X}}}\\
{d_{2,1}}&0&{\cdots}&{d_{2,N_\mathbf{X}}}\\
{\vdots}&{\vdots}& \ddots & \vdots \\
{d_{N_\mathbf{X},1}}&{d_{N_\mathbf{X},2}}& \cdots &0
\end{array}} \right],
\end{equation}
where $d_{ij}$ is the Euclidean distance of points $\mathbf{x}_i, \mathbf{x}_j$, $\mathbf{D} \in \mathbb{R}^{N_\mathbf{X} \times N_\mathbf{X}}$. Analogously, we achieve the edge matrix of the learned RCPs $\mathbf{Y}^{\prime\prime}$ as $\mathbf{D}^{\prime\prime} \in \mathbb{R}^{N_\mathbf{X} \times N_\mathbf{X}}$ . Because $\mathbf{X}$ and $\mathbf{Y}^{\prime\prime}$ are matched sequentially,
we propose to supervise these two edge matrices by defining the loss function as:
\begin{equation}
\mathcal{L}_2 = \text{rmse}(\mathbf{D}, \mathbf{D}^{\prime\prime}).
\label{Eq:double_loss}
\end{equation}

In addition to the edge constraint, we also supervise the node.
Here, we constrain the node distribution by defining the unsupervised loss function as follows:
\begin{equation}
\mathcal{L}_3 =  \text{rmse}(\textbf{R}^\text{pred}\mathbf{X}+\mathbf{t}^\text{pred}, \mathbf{Y}^{\prime\prime}),
\label{Eq:global_loss}
\end{equation}
where $\textbf{R}^\text{pred}$ and $\textbf{t}^\text{pred}$ is the predicted rigid motion. 
\textit{By the geometry structure supervision, we emphasize the geometry structure consistency of the two point clouds.}

\noindent\textbf{Amendment offset supervision.}
In addition to emphasizing the rigidity and geometry structure, a special supervised loss function is proposed here to supervise the amendment offset explicitly, which is defined as:
\begin{equation}
    \mathcal{L}_4 =  \text{rmse}(\textbf{R}^\text{gt}\mathbf{X}+\mathbf{t}^\text{gt}- \mathbf{Y}^{\prime}, \Delta \mathbf{t}_\mathbf{X}),
\label{Eq:finetune_loss}
\end{equation}
where $\textbf{R}^\text{gt}$ and $\textbf{t}^\text{gt}$ represent the ground truth rigid motion. \textit{By amendment offset supervision, we enforce the correction-walk module to learn the offset accurately.}

In our implementation, we train the feature extractor at first by $\mathcal{L}_0$. Then, we freeze this part and train the correction-walk module by $\mathcal{L}_1$, $\mathcal{L}_2$, $\mathcal{L}_3$ and $\mathcal{L}_4$ as follows, where $\lambda_1$, $\lambda_2$, $\lambda_3$ and $\lambda_4$ are trade-off parameters.
\begin{equation}
\mathcal{L} =  \lambda_1 \mathcal{L}_1+ \lambda_2 \mathcal{L}_2+ \lambda_3 \mathcal{L}_3 + \lambda_4 \mathcal{L}_4.
\label{Eq:loss}
\end{equation}
Note that $\mathcal{L}_4$ is a supervised loss function and $\mathcal{L}_1$, $\mathcal{L}_2$, $\mathcal{L}_3$ are unsupervised loss functions.

\subsection{Implementation details}

\noindent\textbf{Network architecture details.} The framework is shown in \figref{Fig:network}, where two deep learning modules exist, \ie feature extractor and correction-walk. The feature extractor module consists of the DGCNN and the Transformer. The correction-walk module is an MLP network.

In DGCNN part, five edge convolution layers are used where the numbers of filters are set to $[64,64,128,256,512]$. In each edge convolution layer, BatchNorm and ReLU are taken. The parameter $K$ in \equref{equ:dgcnn} is set to 20. In Transformer part, one encoder and one decoder are applied with 4 heads and the embedding dimension $c$ is set to 512. 

The output dimension of the correction-walk module is set as $[$512,256,512, 256,128,16,3$]$. Except for the final layer, BatchNorm and ReLU are applied. The final 3D output represents the ``walk'' on the XYZ coordinates.

\noindent\textbf{Training the network.} 
At first, we train the network using $\mathcal{L}_0$ by the Adam optimizer with an initial learning rate of 1e-3 and the batch size of 28 for 100 epochs. We implement the network with PyTorch and train it on GTX 1080Ti. Then, we freeze the feature extractor part and fine-tune the network for 100 epochs resetting the initial learning rate to 1e-4. Here, we use the loss $\mathcal{L}$ and empirically set $\lambda_1=\lambda_2=\lambda_3=1.0$ and $\lambda_4=100$ in \equref{Eq:loss}. We set $G=10$ in \equref{Eq:local_loss}.

%%%%%%%%%%%%%%%%%%%%
\section{Experiments and evaluation}
\label{4_experiments}
%%%%%%%%%%%%%%%%%%%%%%%%%%%%%%%%%%%%%%%%%%%%%%%%%%%%%%%%%%%%%%%%%%%%%%%%%%%%%%%
In this section, to demonstrate the superiority of our proposed method, we perform extensive experiments and comparisons on several benchmark datasets, including the synthetic dataset, ModelNet40 \cite{wu_modelnet40_cvpr_2015}, real indoor dataset, SUN3D \cite{xiao_sun3d_iccv_2013} and 3DMatch \cite{zeng_3dmatch_cvpr_17}, and real outdoor dataset, KITTI \cite{geiger_kitti_rr_13}.

\begin{table*}[ht]
	\renewcommand\arraystretch{1.0}
	\caption{Evaluation on the consistent point clouds. The boldface indicates the best performance and the underline represents the second-best performance. The green indicates the increase of our method compared with the second-best results. And if ours is not the best, it indicates the gap compared with the best results. All tables follow this protocol.}
	\vspace{-0.4cm}
	\begin{center}
		\resizebox{0.8\linewidth}{!}{
			\begin{tabular}{lcccccccccccc}
				\toprule
				\multirow{2}*{\textbf{Method}}&\multicolumn{3}{c}{\textbf{RMSE(R)}}&\multicolumn{3}{c}{\textbf{MAE(R)}}&\multicolumn{3}{c}{\textbf{RMSE(t)}\ ($\times 10^{-4}$)}&\multicolumn{3}{c}{\textbf{MAE(t)}\ ($\times 10^{-4}$)} \\
				
				\cmidrule(r){2-4} \cmidrule(r){5-7} \cmidrule(r){8-10} \cmidrule(r){11-13}
				
				~&\multicolumn{1}{c}{\textit{UPC}}&\multicolumn{1}{c}{\textit{UC}}&\multicolumn{1}{c}{\textit{ND}}&\multicolumn{1}{c}{\textit{UPC}}&\multicolumn{1}{c}{\textit{UC}}&\multicolumn{1}{c}{\textit{ND}}&\multicolumn{1}{c}{\textit{UPC}}&\multicolumn{1}{c}{\textit{UC}}&\multicolumn{1}{c}{\textit{ND}}&\multicolumn{1}{c}{\textit{UPC}}&\multicolumn{1}{c}{\textit{UC}}&\multicolumn{1}{c}{\textit{ND}} \\
				
				\midrule
				\textbf{ICP}       &12.282 &12.707 &11.971 &4.613 &5.075 &4.497 &477.44 &485.32 &483.20 & 22.80 &23.55 & \underline{43.35}\\
				\textbf{FGR}       &20.054 &21.323 &18.359 &7.146 &8.077 &6.367 &441.21 &457.77 &391.01 &164.20 &\underline{18.07} &144.87\\
				\textbf{PTLK}      &13.751 &15.901 &15.692 &3.893 &4.032 &3.992 &199.00 &261.15 &239.58 & 44.52 &62.13 & 56.37\\
				\textbf{DCP-v2}    & \underline{1.094} & 3.256 & 8.417 &0.752 &2.102 &5.685 &\underline{17.17}  & \underline{63.17} &318.37 & \underline{11.73} &46.29 &233.70\\
				\textbf{PRNet}     & 1.722 & \underline{3.060} & \underline{3.218} &\underline{0.665} &\underline{1.326} &\underline{1.446} &63.72  &100.95 &\underline{111.78} & 46.52 &75.89 & 83.78\\
				\midrule
				\textbf{VRNet}  
				&\makecell[c]{\textbf{0.091}\\\scriptsize{\color{SpringGreen}+91.68\%}}
				&\makecell[c]{\textbf{0.209}\\\scriptsize{\color{SpringGreen}+93.17\%}}
				&\makecell[c]{\textbf{2.558}\\\scriptsize{\color{SpringGreen}+20.51\%}}
				&\makecell[c]{\textbf{0.012}\\\scriptsize{\color{SpringGreen}+98.20\%}}
				&\makecell[c]{\textbf{0.028}\\\scriptsize{\color{SpringGreen}+97.89\%}}
				&\makecell[c]{\textbf{1.016}\\\scriptsize{\color{SpringGreen}+29.74\%}}
				&\makecell[c]{\textbf{2.97} \\\scriptsize{\color{SpringGreen}+82.70\%}}
				&\makecell[c]{\textbf{7.83} \\\scriptsize{\color{SpringGreen}+87.60\%}}
				&\makecell[c]{\textbf{57.02}\\\scriptsize{\color{SpringGreen}+48.99\%}}
				&\makecell[c]{\textbf{0.47} \\\scriptsize{\color{SpringGreen}+95.99\%}}
				&\makecell[c]{\textbf{0.99} \\\scriptsize{\color{SpringGreen}+94.52\%}}
				&\makecell[c]{\textbf{28.97}\\\scriptsize{\color{SpringGreen}+33.17\%}} \\
				\bottomrule
			\end{tabular}}
		    \label{Tab:consistent_registration}
	\end{center}
	\vspace{-0.4cm}
\end{table*}

\begin{table*}[t]
	\renewcommand\arraystretch{1.0}
	\caption{Evaluation on point clouds processed by partial-view, random sample, partial-view \& random sample strategy.}
	\vspace{-0.4cm}
	\begin{center}
		\resizebox{0.8\linewidth}{!}{
			\begin{tabular}{lccccccccccccc}
				\toprule
				\multirow{2}*{\textbf{Method}}&\multicolumn{3}{c}{\textbf{RMSE(R)}}&\multicolumn{3}{c}{\textbf{MAE(R)}}&\multicolumn{3}{c}{\textbf{RMSE(t)}}&\multicolumn{3}{c}{\textbf{MAE(t)}} &\\
				
				\cmidrule(r){2-4} \cmidrule(r){5-7} \cmidrule(r){8-10} \cmidrule(r){11-13}
				
				~&\multicolumn{1}{c}{\textit{UPC}}&\multicolumn{1}{c}{\textit{UC}}&\multicolumn{1}{c}{\textit{ND}}&\multicolumn{1}{c}{\textit{UPC}}&\multicolumn{1}{c}{\textit{UC}}&\multicolumn{1}{c}{\textit{ND}}&\multicolumn{1}{c}{\textit{UPC}}&\multicolumn{1}{c}{\textit{UC}}&\multicolumn{1}{c}{\textit{ND}}&\multicolumn{1}{c}{\textit{UPC}}&\multicolumn{1}{c}{\textit{UC}}&\multicolumn{1}{c}{\textit{ND}} &\\
				
				\midrule
				\textbf{ICP}       &33.683 &34.894 &35.067 &25.045 &25.455 &25.564 &0.293 &0.293 &0.294 &0.250 &0.251 &0.250 &\multirow{7}{*}{\rotatebox{90}{PV}}\\
				\textbf{FGR}       &11.238 & 9.932 &27.653 &2.832  &\underline{1.952}  &13.794 &0.030 &0.038 &0.070 &\underline{0.008} &\underline{0.007} &0.039&\\
				\textbf{PTLK}      &16.735 &22.943 &19.939 &7.550  &9.655  &9.076  &0.045 &0.061 &0.057 &0.025 &0.033 &0.032&\\
				\textbf{DCP-v2}    & 6.709 & 9.769 & 6.883 &4.448  &6.954  &4.534  &0.027 &0.034 &0.028 &0.020 &0.025 &0.021&\\
				\textbf{PRNet}     & \underline{3.199} & \underline{4.986} & \underline{4.323} &\underline{1.454}  &2.329  &\underline{2.051}  &\underline{0.016} &\underline{0.021} &\underline{0.017} &0.010 &0.015 &\underline{0.012}&\\
				\cmidrule(r){1-13}
				\textbf{VRNet}  
				&\makecell[c]{\textbf{0.982}\\\scriptsize{\color{SpringGreen}+69.30\%}}
				&\makecell[c]{\textbf{2.121}\\\scriptsize{\color{SpringGreen}+57.46\%}}
				&\makecell[c]{\textbf{3.615}\\\scriptsize{\color{SpringGreen}+16.38\%}}
				&\makecell[c]{\textbf{0.496}\\\scriptsize{\color{SpringGreen}+65.89\%}}
				&\makecell[c]{\textbf{0.585}\\\scriptsize{\color{SpringGreen}+70.03\%}}
				&\makecell[c]{\textbf{1.637}\\\scriptsize{\color{SpringGreen}+20.19\%}} 
				&\makecell[c]{\textbf{0.0061}\\\scriptsize{\color{SpringGreen}+61.88\%}}
				&\makecell[c]{\textbf{0.0063}\\\scriptsize{\color{SpringGreen}+70.00\%}}
				&\makecell[c]{\textbf{0.0101}\\\scriptsize{\color{SpringGreen}+40.59\%}} 
				&\makecell[c]{\textbf{0.0039}\\\scriptsize{\color{SpringGreen}+51.25\%}}
				&\makecell[c]{\textbf{0.0039}\\\scriptsize{\color{SpringGreen}+44.29\%}}
				&\makecell[c]{\textbf{0.0063}\\\scriptsize{\color{SpringGreen}+47.50\%}}& \\
				\midrule \midrule
				\textbf{ICP}       &11.247 &12.723 &11.472 &4.531  &5.289  &4.752  &0.0421 &0.0454 &0.0430   &0.0232 &0.0231 &0.0249&\multirow{7}{*}{\rotatebox{90}{RS}}\\
				\textbf{FGR}       &19.293 &19.62  &37.452 &7.054  &7.566  &23.230 &0.0414 &0.0433 &360.6727 &0.0157 &0.0167 &6.0463&\\
				\textbf{DCP-v2}    & 5.018 & 6.015 & 5.536 &2.921  &3.964  &3.162  &\underline{0.0116} &0.0147 &\underline{0.0127}   &\underline{0.0087} &0.0112 &\underline{0.0096}&\\
				\textbf{PRNet}     & \underline{4.851} & \textbf{3.484} & \underline{3.824} &\underline{2.429}  &\textbf{1.764}  &\underline{1.781}  &0.0174 &\underline{0.0129} &0.0128 &0.0134 &\underline{0.0100} &0.0099&\\
				\cmidrule(r){1-13}
				\textbf{VRNet}  
				&\makecell[c]{\textbf{1.496}\\\scriptsize{\color{SpringGreen}+69.16\%}}
				&\makecell[c]{\underline{5.651}\\\scriptsize{\color{SpringGreen}-62.20\%}}
				&\makecell[c]{\textbf{3.099}\\\scriptsize{\color{SpringGreen}+18.96\%}} 
				&\makecell[c]{\textbf{0.593}\\\scriptsize{\color{SpringGreen}+75.59\%}}
				&\makecell[c]{\underline{1.971}\\\scriptsize{\color{SpringGreen}-11.73\%}}
				&\makecell[c]{\textbf{1.476}\\\scriptsize{\color{SpringGreen}+17.13\%}} 
				&\makecell[c]{\textbf{0.0025}\\\scriptsize{\color{SpringGreen}+78.45\%}}
				&\makecell[c]{\textbf{0.0041}\\\scriptsize{\color{SpringGreen}+68.22\%}}
				&\makecell[c]{\textbf{0.0077}\\\scriptsize{\color{SpringGreen}+39.37\%}} 
				&\makecell[c]{\textbf{0.0016}\\\scriptsize{\color{SpringGreen}+81.61\%}}
				&\makecell[c]{\textbf{0.0071}\\\scriptsize{\color{SpringGreen}+36.61\%}}
				&\makecell[c]{\textbf{0.0057}\\\scriptsize{\color{SpringGreen}+40.63\%}}& \\
				\midrule \midrule
				\textbf{ICP}       &11.971 &13.669 &12.215 &5.298  &6.018  &5.624  &0.0499 &0.0521 &0.0502   &0.0296 &0.0287 &0.0309&\multirow{7}{*}{\rotatebox{90}{PV \& RS}}\\
				\textbf{FGR}       &7.837  &9.187  &32.491 &\underline{2.076}  &2.017  &14.680 &0.0167 &0.0168 &0.0604   &\underline{0.0063} &\underline{0.0065} &0.0360&\\
				\textbf{DCP-v2}    & 5.818 &7.059  & 6.286 &3.399  &4.564  &3.844  &0.0184 &0.0176 &0.0222   &0.0138 &0.0131 &0.0168&\\
				\textbf{PRNet}     & \underline{4.924} & \underline{3.836} & \underline{4.519} &2.573  &\underline{1.901}  &\underline{1.925}  &\underline{0.0162} &\underline{0.0163} &\underline{0.0139}   &0.0112 &0.0112 &\underline{0.0099}&\\
				\cmidrule(r){1-13}
				\textbf{VRNet}  
				&\makecell[c]{\textbf{1.109}\\\scriptsize{\color{SpringGreen}+77.48\%}}
				&\makecell[c]{\textbf{1.842}\\\scriptsize{\color{SpringGreen}+51.98\%}}
				&\makecell[c]{\textbf{2.411}\\\scriptsize{\color{SpringGreen}+46.65\%}} 
				&\makecell[c]{\textbf{0.513}\\\scriptsize{\color{SpringGreen}+75.29\%}}
				&\makecell[c]{\textbf{0.702}\\\scriptsize{\color{SpringGreen}+63.07\%}}
				&\makecell[c]{\textbf{1.020}\\\scriptsize{\color{SpringGreen}+47.01\%}} 
				&\makecell[c]{\textbf{0.0037}\\\scriptsize{\color{SpringGreen}+77.16\%}}
				&\makecell[c]{\textbf{0.0045}\\\scriptsize{\color{SpringGreen}+72.39\%}}
				&\makecell[c]{\textbf{0.0072}\\\scriptsize{\color{SpringGreen}+48.20\%}} 
				&\makecell[c]{\textbf{0.0023}\\\scriptsize{\color{SpringGreen}+63.49\%}}
				&\makecell[c]{\textbf{0.0026}\\\scriptsize{\color{SpringGreen}+60.00\%}}
				&\makecell[c]{\textbf{0.0050}\\\scriptsize{\color{SpringGreen}+49.49\%}}& \\
				\bottomrule
			\end{tabular}}
	    \label{Tab:partial}
	\end{center}
    \vspace{-0.4cm}
\end{table*}

%-------------------------------------------------------------
\subsection{Evaluation on synthetic dataset: ModelNet40} \label{sec:exp:modelnet40}
%%%%%%%%%%%%%%%%%%%%%%%%%%%%%%%%%%%%%%%%%%%%%%%%%%%%%%%%%%%%%%%

%\vspace{0.2cm}
\noindent\textbf{Dataset and processing.}
We first evaluate our method on ModelNet40 \cite{wu_modelnet40_cvpr_2015}, which is a synthetic dataset consisting of 3D CAD models from 40 categories. ModelNet40 is a widely used benchmark in point cloud registration evaluation \cite{aoki_ptlk_cvpr_2019,wang_dcp_iccv_2019,wang_prnet_nips_2019,yew_rpmnet_cvpr_2020,huang_featuremetric_cvpr_2020}. Following \cite{wang_dcp_iccv_2019,wang_prnet_nips_2019,yew_rpmnet_cvpr_2020}, we sample 1024 points randomly as the \textit{source} point cloud $\mathbf X$. And then, $\mathbf{X}$ is rigidly transformed to generate the \textit{target} point cloud $\mathbf{Y}$. Because the dataset is synthetic and the \textit{target} is generated from the \textit{source}, the correspondences are obtained naturally. The employed rotation and translation are uniformly sampled in $[0^ \circ, 45^ \circ]$ and $[-0.5, 0.5]$ respectively along each axis. Both the \textit{source} and \textit{target} point clouds are shuffled. These settings are widely adopted in the community for a fair comparison.

We test the proposed VRNet on ModelNet40 with/without outliers. Here, four data processing settings are provided as follows. Note that each of the consistent input point clouds consists of 1024 points as mentioned above.
\begin{itemize}
\setlength{\itemsep}{0pt}
\setlength{\parsep}{0pt}
\setlength{\parskip}{0pt}
    \item \underline{Co}nsistent point clouds (\textbf{\textit{CO}}). The \textit{source} and \textit{target} point clouds are exactly same except for the pose, \ie each point has a corresponding point in the paired point cloud. %In this setting, there are no outliers exist in the \textit{source} and \textit{target}.
    \item \underline{P}artial-\underline{v}iew (\textbf{\textit{PV}}). Following PRNet \cite{wang_prnet_nips_2019}, given a random point in 3D space, we select its nearest 768 points from the original consistent input point clouds. However, despite this strategy is widely adopted, it results in the same distribution of overlapped parts, and limited partiality is obtained with low outliers ratio.
    \item \underline{R}andom-\underline{s}ample (\textbf{\textit{RS}}). We randomly select 768 points from each consistent point cloud, leading to the random distribution of outliers and high outliers ratio.
    \item \underline{P}artial-\underline{v}iew \& \underline{R}andom-\underline{s}ample (\textbf{\textit{PV+RS}}). A more challenging data processing setting is provided by combining the above partial-view and random sample strategies. Specifically, we select 896 points randomly from each consistent point cloud at first, and then 768 nearest points are selected from these sampled 896 points, respectively.
\end{itemize}

\noindent\textbf{Dataset split setting.}
Following \cite{wang_dcp_iccv_2019,wang_prnet_nips_2019}, three dataset split settings are applied here for a comprehensive evaluation.

\begin{itemize}
\setlength{\itemsep}{0pt}
\setlength{\parsep}{0pt}
\setlength{\parskip}{0pt}
	\item {\underline{U}nseen \underline{P}oint \underline{C}louds} (\textbf{\textit{UPC}}). The ModelNet40 is divided into training and testing sets with the official setting.
	\item {\underline{U}nseen \underline{C}ategories} (\textbf{\textit{UC}}). To test the generalization ability to the unseen point cloud categories, we divide ModelNet40 according to the object category. The first 20 categories are selected for training and the rest are used for testing. This setting is consistent with \cite{wang_dcp_iccv_2019, wang_prnet_nips_2019, yew_rpmnet_cvpr_2020}.
	\item {\underline{N}oisy \underline{D}ata} (\textbf{\textit{ND}}). To test the robustness, random Gaussian noise (\ie, $\mathcal{N}(0,0.01)$, and the sampled noise out of the range of $[-0.05,0.05]$ will be clipped) is added to each point. The dataset splitting is the same as \textit{UPC}. 
\end{itemize}

\noindent\textbf{Evaluation metrics.}
Following \cite{wang_dcp_iccv_2019, wang_prnet_nips_2019}, root mean square error (RMSE), and mean absolute error (MAE) between the ground truth and the prediction in Euler angle and translation vector are used as our evaluation metrics, notated as {RMSE(R)}, {MAE(R)}, {RMSE(t)} and {MAE(t)} respectively. 

\noindent\textbf{Performance evaluation.} Herein, we present the rigid motion estimation results in the mentioned settings for a comprehensive comparison. Meanwhile, we also provide the time-efficiency to validate the proposed VRNet.

$\bullet$ \textbf{Consistent point clouds:}
Following the protocol of DCP \cite{wang_dcp_iccv_2019}, we take consistent point clouds as our input. The results are reported in \tabref{Tab:consistent_registration}. 
In \textbf{\textit{UPC}} setting, among all baselines, PRNet \cite{wang_prnet_nips_2019} achieves the best performance in MAE(R) and DCP-v2 obtains the best results in RMSE(R), RMSE(t) and MAE(t). However, our proposed VRNet are better than all these baselines. 
In \textbf{\textit{UC}} setting, VRNet maintains the best performance while PRNet achieves the second-best results for rotation estimation, and DCP-v2, FGR obtain the second-best results in RMSE(t) and MAE(t) respectively.
In \textbf{\textit{ND}} setting, our proposed VRNet improves the performance to a large extent in all evaluation metrics compared with all baselines, which validates the robustness of our method further.

%\vspace{0.2cm}
$\bullet$ \textbf{Partial-view:} Following PRNet \cite{wang_prnet_nips_2019}, we test the performance of the proposed method using the partial-view input point clouds. We report the registration results in \tabref{Tab:partial}, where the proposed method achieves the best results in all evaluation metrics including RMSE(R) MAE(R), RMSE(t) and MAE(t) in all dataset split settings including \textbf{\textit{UPC}}, \textbf{\textit{UC}} and \textbf{\textit{ND}}. Besides, PRNet, which is designed for partial-view setting specifically, achieves the second-best performance in most evaluations.

$\bullet$ \textbf{Random-sample:} Due to the random sample strategy, the outliers distribute randomly. The registration performance is reported in \tabref{Tab:partial}. Our VRNet achieves the best performance in \textbf{\textit{UPC}}, \textbf{\textit{ND}}. In \textbf{\textit{UC}}, VRNet obtains the best estimation of translation, and the second-best rotation estimation results.

$\bullet$ \textbf{Partial-view \& Random-sample:} In this part, we combine the above two data processing strategies to evaluate our VRNet, and the results are presented in \tabref{Tab:partial}. Obviously, our VRNet achieves the best performance in all evaluation metrics including RMSE(R), MAE(R), RMSE(t) and MAE(t) in all dataset split settings including \textbf{\textit{UPC}}, \textbf{\textit{UC}}, and \textbf{\textit{ND}}.

$\bullet$ \textbf{Time-efficiency:}
We counts the average inference time of all learning-based methods in \textbf{partial-view} setting using a Xeon E5-2640 v4@2.40GHz CPU and a GTX 1080 Ti, where 768 points exist in each input point cloud. \tabref{Tab:times} shows that ours achieves advanced time-efficiency meeting the real-time requirement since the complicated processes of inliers recognition and reliable correspondences selection are avoided.

\begin{table}[!h]
	\renewcommand\arraystretch{1.0}
	\caption{Inference time comparison on ModelNet40.}	
	\vspace{-0.2cm}
	\centering
	\resizebox{0.8\linewidth}{!}{
		\begin{tabular}{lcccccc}
			\toprule
			\textbf{Methods} &  \textbf{PTLK} &\textbf{DCP} & \textbf{PRNet} & \textbf{RPMNet} & \textbf{VRNet} \\
			\midrule				
			\textbf{Time[ms]}  &47.2 &\textbf{17.66} & 37.48 & 54.75 & \underline{19.92}\\
			\bottomrule
	\end{tabular}}
	\label{Tab:times}
	\vspace{-0.5cm}
\end{table}

%----------------------------------------------------------------
\subsection{Evaluation on real indoor dataset: SUN3D, 3DMatch} \label{sec:exp:indoor}
%%%%%%%%%%%%%%%%%%%%%%%%%%%%%%%%%%%%%%%%%%%%%%%%%%%%%%%%%%%%%%%%
\noindent\textbf{Dataset.} Besides the synthetic dataset, we also conduct evaluations on real indoor scene dataset, SUN3D \cite{xiao_sun3d_iccv_2013} and 3DMatch \cite{zeng_3dmatch_cvpr_17}. SUN3D is composed of 13 randomly selected scenes, and the version processed by 3DRegNet \cite{pais_3dregnet_cvpr_2020} is used here, which is a sparse dataset with around 3000 points in each point cloud. 3DMatch is a hybrid indoor dataset, and the input has been voxelized with the voxel size of 5cm following \cite{choy_dgr_cvpr_2020}. This dataset is a dense, large-scale dataset and each point cloud contains around 50000 points. To train the network, we construct the ground truth correspondences manually. Specifically, we transform the \textit{source} point cloud based on the ground truth transformation matrix first. Then, the nearest neighbor searching is applied to solve the corresponding points. Notably, if the distance between the transformed point and the searched corresponding point is greater than a predefined threshold (\eg 3cm for SUN3D and 5cm for 3DMatch in our implementation), the match is discarded.

\noindent\textbf{Evaluation metrics.} For a fair comparison, we follow the evaluation metrics of 3DRegNet \cite{pais_3dregnet_cvpr_2020} and DGR \cite{choy_dgr_cvpr_2020} respectively.
For SUN3D, we report the mean, median \underline{r}otation \underline{e}rror (RE), and the mean, median \underline{t}ranslation \underline{e}rror (TE), and time-efficiency (Time), where RE and TE are calculated by
\begin{equation}
    \left\{
    \begin{aligned}
    \text{RE}&=\text{arccos}(({\textrm{trace}(\mathbf{R}^{-1}\mathbf{R}^\text{gt})-1)/2})\frac{180}{\pi} \\
    \text{TE}&=\|\mathbf{t}-\mathbf{t}^\text{gt}\|_2^2
    \end{aligned}
    \right. ,
\end{equation}
where the superscript ``$\text{gt}$'' indicates the ground truth. For 3DMatch, besides the mean RE, mean TE and time-efficiency, we also report the recall following \cite{choy_dgr_cvpr_2020}, which is the ratio of successful pairwise registrations. Here, the successful pairwise is confirmed if the rotation error and translation error are smaller than the pre-defined thresholds (\ie 15 deg, 30cm). It is worth mentioning that, mean RE and mean TE reported in \tabref{Tab:3dmatch} are computed only on these successfully registered pairs since the relative poses returned from the failed registrations can be drastically different from the ground truth, making the error metrics unreliable.

\begin{table}[h]
	\renewcommand\arraystretch{1.0}
	\caption{Comparison of registration results in SUN3D following the metrics of 3DRegNet \cite{pais_3dregnet_cvpr_2020}}
	\vspace{-0.4cm}
    \begin{center}
	    \resizebox{0.9\linewidth}{!}{
			\begin{tabular}{lccccc}
				\toprule
				\multirow{2}*{\textbf{Methods}}&\multicolumn{1}{c}{\textbf{RE}[deg]}
				                               &\multicolumn{1}{c}{\textbf{RE}[deg]}
				                               &\multicolumn{1}{c}{\textbf{TE}[m]} 
				                               &\multicolumn{1}{c}{\textbf{TE}[m]}
				                               &\multirow{2}*{\textbf{Time}[ms]}\\
				\cmidrule(r){2-3} \cmidrule(r){4-5} 
				~&\textbf{Mean}&\textbf{Median}&\textbf{Mean}&\textbf{Median}\\
				\midrule				
				\textbf{FGR}    & 2.57  & 1.92              & 0.121 & \underline{0.067} &44.34\\
				\textbf{ICP}    & 3.18  & \underline{1.50}  & 0.146 & 0.079             &\underline{32.17}\\
				\textbf{RANSAC} & 3.00  &1.73               & 0.148 & 0.074             &170.2\\
				\midrule
				\textbf{3DRegNet} &\underline{1.84} &1.69   &\underline{0.087} &0.078   &166.7\\
				\midrule
				\textbf{VRNet}
				&\makecell[c]{\textbf{1.49}\\\scriptsize{\color{SpringGreen}+19.02\%}}
				&\makecell[c]{\textbf{0.38}\\\scriptsize{\color{SpringGreen}+74.67\%}}
				&\makecell[c]{\textbf{0.075}\\\scriptsize{\color{SpringGreen}+13.79\%}}
				&\makecell[c]{\textbf{0.058}\\\scriptsize{\color{SpringGreen}+13.43\%}} &\textbf{25.6}\\
				\bottomrule
		\end{tabular}}
		\label{Tab:sun3d_regnet}
 	\vspace{-0.2cm}
    \end{center}
\end{table}

\noindent\textbf{Performance evaluation.} 
In \tabref{Tab:sun3d_regnet}, we provide the performance comparison on SUN3D. Traditional methods, including ICP \cite{besl_icp_pami_1992}, FGR \cite{zhou_fgr_eccv_2016} and RANSAC, all present an acceptable registration result. Moreover, ICP and FGR even achieve the second-best performance on median TE and median RE metrics respectively within all methods. The learning-based method, 3DRegNet \cite{pais_3dregnet_cvpr_2020} presents the better performance on mean RE and mean TE than these traditional methods, however, it is complicated. We also conduct DCP-v2 and PRNet, unfortunately, they fail in this setting with divergent results. Our proposed VRNet outperforms all these baselines on both transformation estimation and time-efficiency performance, which validates the superiority of our method.

\begin{table}[h]
	\renewcommand\arraystretch{1.0}
	\caption{Evaluation on 3DMatch dataset.}	
	\vspace{-0.4cm}
	\begin{center}
	\resizebox{0.9\linewidth}{!}{
		\begin{tabular}{lcccc}
			\toprule
				\textbf{Methods} & \textbf{TE}[cm] & \textbf{RE}[deg] & \textbf{Recall}(\%) & \textbf{Times}[s]\\
				\midrule		
				\textbf{ICP}                 & 18.1    & 8.25   & 6.04   & 0.25   \\
				\textbf{FGR}                 & 10.6    & 4.08   & 42.7   & 0.31   \\
				\textbf{Go-ICP}              & 14.7    & 5.38   & 22.9   & 771.0  \\
				\textbf{Super4PCS}           & 14.1    & 5.25   & 21.6   & 4.55   \\
				\textbf{RANSAC}              & 8.85    & \underline{3.00}   & 66.1   & 1.39   \\
				\midrule
				\textbf{DCP-v2}                 & 21.4    & 8.42   & 3.22   & \textbf{0.07}   \\
				\textbf{PTLK}                & 21.3    & 8.04   & 1.61   & 0.12   \\
				\textbf{DGR}                 & \textbf{7.34}    & \textbf{2.43}   & \textbf{91.3}   & 1.21   \\
				\midrule
				\textbf{VRNet}               & \underline{8.64}    & 3.12   & \underline{72.9}   & \underline{0.11}\\
				\bottomrule
	\end{tabular}}
	\label{Tab:3dmatch}
	\end{center}
	\vspace{-0.2cm}
\end{table}

\tabref{Tab:3dmatch} presents the evaluation results on 3DMatch, we find that ICP \cite{besl_icp_pami_1992} achieves the weakest performance because the dataset is so challenging since the large rigid motion exists and there is no reliable prior provided. Meanwhile, a sampling-based algorithm, Super4PCS  \cite{mellado_super4pcs_cgf_14}, and the ICP variant with branch-and-bound search, Go-ICP \cite{yang_goicp_pami_2015} perform similarly. The feature-based methods, \ie FGR and RANSAC, perform much better than these methods which are built on the 3D point directly. 
As for learning-based methods, DGR \cite{choy_dgr_cvpr_2020}, which is designed for dense scene dataset specifically, achieves the best performance in all registration metrics. However, because DGR devotes to selecting reliable correspondence, it is complicated and time-consuming. PointNetLK is a correspondences-free method, which fails in this setting because of the numerous outliers. DCP-v2 also fails here despite the best time-efficiency is achieved, we suspect that the feature extractor of DCP-v2 is not suitable to 3DMatch dataset. Inspired by this, we use the FCGF \cite{choy_fcgf_iccv_2019} feature extractor instead in our VRNet, which is designed for such large-scale scene point clouds specifically. Furthermore, although only the second-best rigid transformation estimation results are achieved by ours, which is weaker than DGR, our VRNet obtains the better time-efficiency when ignoring the failed method, DCP. This characteristic of balancing the transformation estimation and the running time is crucial in actual applications.

%------------------------------------------------
\subsection{Evaluation on real outdoor data: KITTI} \label{sec:exp:kitti}

The typical outdoor scene dataset, KITTI \cite{geiger_kitti_rr_13} is used here to evaluate our VRNet, which consists of LIDAR scans. Following \cite{choy_dgr_cvpr_2020}, we build the point cloud pairs with at least 10m apart, and the ground-truth transformation is generated using GPS followed by ICP to fix the inherent errors. The strategy to construct ground truth correspondences is the same as the strategy for the SUN3D, 3DMatch datasets, and the threshold to determine the acceptable correspondences is set to 5cm.
Besides, we use the voxel size of 30cm to downsample the input point clouds. It is worth mentioning that we use the FCGF \cite{choy_fcgf_iccv_2019} feature extractor in our VRNet as \secref{sec:exp:indoor}, which is designed for such large-scale scene dataset.  
\tabref{Tab:kitti} reports the registration performance. In learning-based approaches, the proposed VRNet obtains better performance on translation estimation while DGR \cite{choy_dgr_cvpr_2020} achieves better rotation estimation results. Meanwhile, the best time-efficiency is achieved by our VRNet, which is important in actual applications. DCP \cite{wang_dcp_iccv_2019} and PRNet \cite{wang_prnet_nips_2019} also can make sense in this setting. However, since the GPU consuming is out of memory for ``DGCNN + transformer'' feature extractor in DCP and PRNet, we have to downsample the input point clouds with the voxel size of 70cm further for these two baselines. Thus, we do not report their time-efficiency here for a fair comparison.

\begin{table}[h]
	\renewcommand\arraystretch{1.0}
	\caption{Evaluation on KITTI dataset.}	
	\vspace{-0.4cm}
	\begin{center}
	\resizebox{0.9\linewidth}{!}{
		\begin{tabular}{lccccc}
			\toprule
			    \multirow{2}*{\textbf{Methods}}&\multicolumn{2}{c}{\textbf{Rotation}[deg]}
				                               &\multicolumn{2}{c}{\textbf{Translation}[cm]} 
				                               &\multirow{2}*{\textbf{Time}[s]}
				 \\
				 \cmidrule(r){2-3} \cmidrule(r){4-5} 
				~&\textbf{RMSE}&\textbf{MAE}&\textbf{RMSE}&\textbf{MAE}\\
				\midrule		
				\textbf{ICP}   & \underline{7.54}          & \textbf{2.10} & 4.84            & 2.94          & \underline{1.44} \\
				\textbf{FGR}   & 60.45         & 27.81         &42.1             &12.69          & 1.50 \\
				\midrule
				\textbf{DCP}   & 11.09         & 10.42         &18.04            &16.29          & - \\%$\ast$\\
				\textbf{PRNet} & 10.93         & 7.86          &13.28            &10.52          & - \\%$\ast$\\
				\textbf{DGR}   & \textbf{5.59} & \underline{2.32}          &\underline{4.72}             &\underline{2.31}           & 2.42 \\
				\midrule
				\textbf{VRNet} & 7.56          & 3.42          &\textbf{1.72}     &\textbf{1.18} & \textbf{0.24}\\
				\bottomrule
	\end{tabular}}
	\label{Tab:kitti}
	\end{center}
	\vspace{-0.6cm}
\end{table}

%%%%%%%%%%%%%%%%%%%%%%%%%%%%%%%%%%%%%%%%%%%%%%%%%
\subsection{Ablation studies} \label{4-5-ablation}
%%%%%%%%%%%%%%%%%%%%%%%%%%%%%%%%%%%%%%%%%%%%%%%%%

\noindent\textbf{The consistency comparison.}
The essence of our method is to construct correspondences for all \textit{source} points without distinguishing outliers and inliers. To this end, we drive the learned corresponding points to keep rigidity and geometry structure consistency with the \textit{source} point cloud. Here, we measure this consistency using the Chamfer distance, which can be calculated as follows, 
\begin{equation}
\textbf{CD}(\mathbf{X},\mathbf{Y}) = 
 \frac{1}{N_\mathbf{X}} \sum \limits_{\mathbf{x} \in \mathbf{X}} \min \limits_{\mathbf{y}\in \mathbf{Y}} \|\mathbf{x}-\mathbf{y}\|_2^2
 + \frac{1}{N_\mathbf{Y}} \sum \limits_{\mathbf{y}\in \mathbf{Y}} \min \limits_{\mathbf{x}\in \mathbf{X}} \|\mathbf{x}-\mathbf{y}\|_2^2.
\end{equation}

\begin{figure}[h]
    \centering\includegraphics[width=0.9\linewidth]{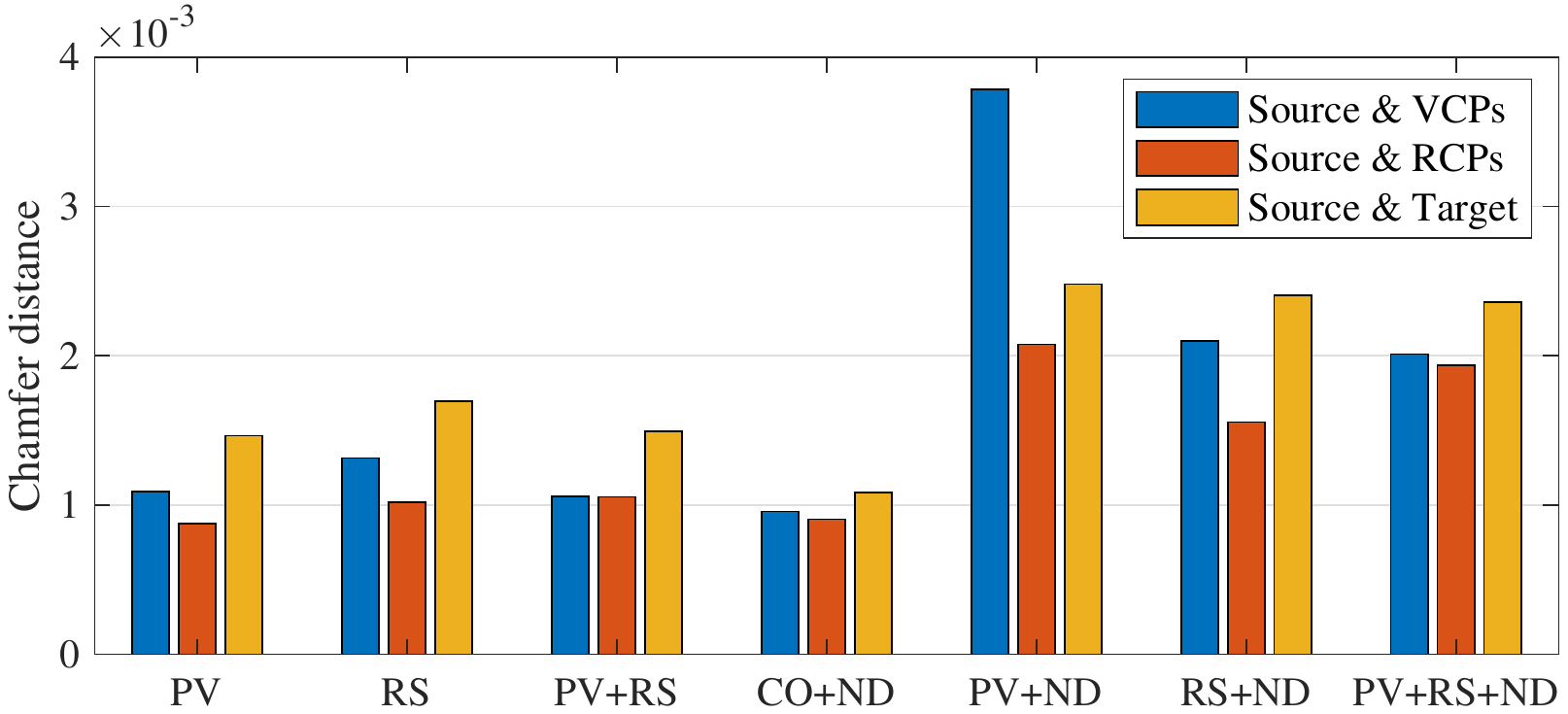}
	\caption{We count the Chamfer distance in different settings including \textit{partial-view}, \textit{random-sample}, \textit{partial-view}\&\textit{random-sample}, \textit{consistent} point clouds in \textit{noise} ($CO+ND$), \textit{partial-view} in \textit{noise} ($PV+ND$), \textit{random-sample} in \textit{noise} ($RS+ND$), \textit{partial-view}\&\textit{random-sample} in \textit{noise} ($PV+RS+ND$).
	The corresponding points learned by the VRNet are more consistent to the \textit{source}, because the Chamfer distances between the \textit{source} and the learned RCPs  are always smallest.}
	\label{Fig:CD}
\end{figure}

We provide the comparison in \figref{Fig:CD}, where ``Source \& VCPs'' represents the Chamfer distance between the \textit{source} point cloud and the learned VCPs, ``Source \& RCPs'' represents the Chamfer distance between the \textit{source} and the learned RCPs, and ``Source \& Target'' represents the Chamfer distance between the \textit{source} and \textit{target}.
The \textit{source} has been transformed with the ground truth rigid transformation. 
As shown in \figref{Fig:CD}, because of outliers, the \textit{source} and the \textit{target} are not entirely the same, thus even the ground truth transformation is applied, the corresponding Chamfer distance is still large. Meanwhile, because the distribution limitation has been broken, the learned RCPs are more consistent with the \textit{source} point cloud in which the less Chamfer distance is obtained.

\begin{figure*}[h]
	\centerline{\includegraphics[width=1.0\linewidth]{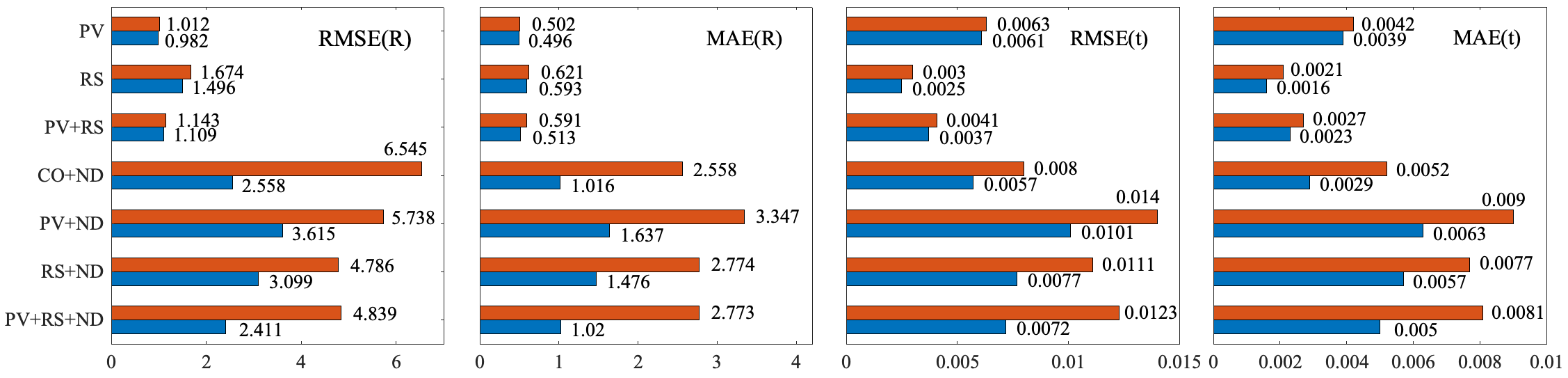}}
	\caption{We show the improvement caused by the correction-walk module. The orange/blue indicates the results without/with the correction-walk module. In all settings and all metrics, the registration results are improved.}
	\label{Fig:improve}
\end{figure*}

\noindent\textbf{Registration improvement by the correction-walk module.}
To break the distribution limitation, we propose a correction-walk module to learn an offset to amend the corresponding points. Herein, we give a quantitative effectiveness analysis of this module by comparing the registration performance. As shown in \figref{Fig:improve}, via our correction-walk module, the registration performance shows an obvious improvement in all metrics and all settings.

\begin{figure*}[h]
    \centering
    \includegraphics[width=0.8\linewidth]{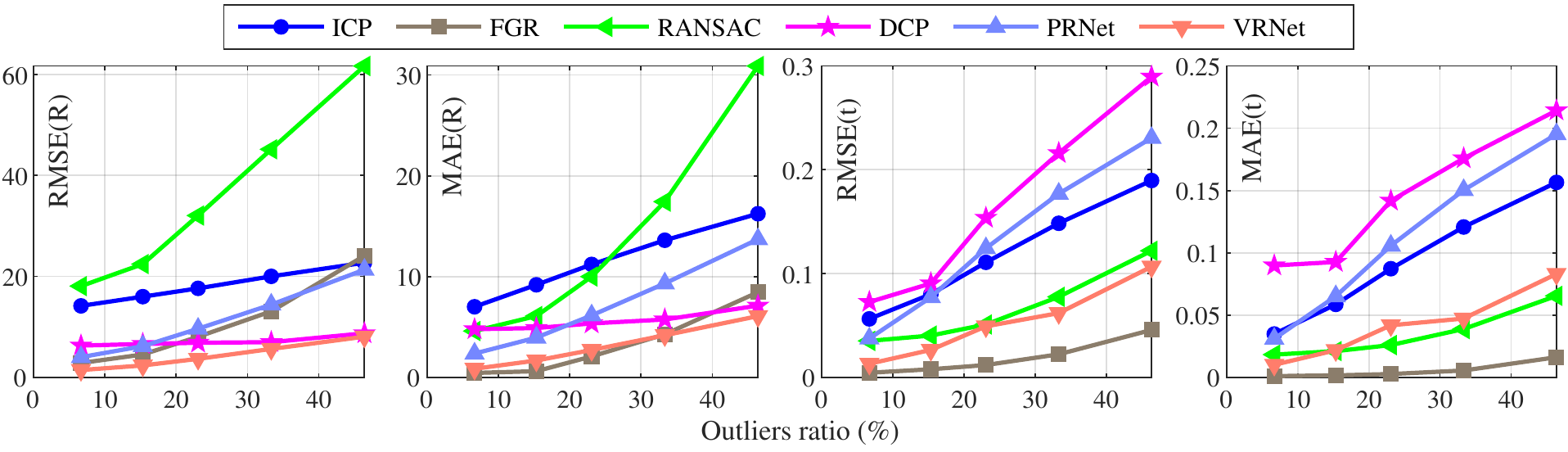}
    \caption{The registration performance decreases as the outliers ratio increases. Traditional methods, like RANSAC, FGR, present excellent performance. Our VRNet achieves state-of-the-art performance in rotation estimation and the best translation estimation in learning-based methods.}
    \label{Fig:degeneration}
\end{figure*}

\begin{figure*}[ht]
    \centering
    \includegraphics[width=0.8\linewidth]{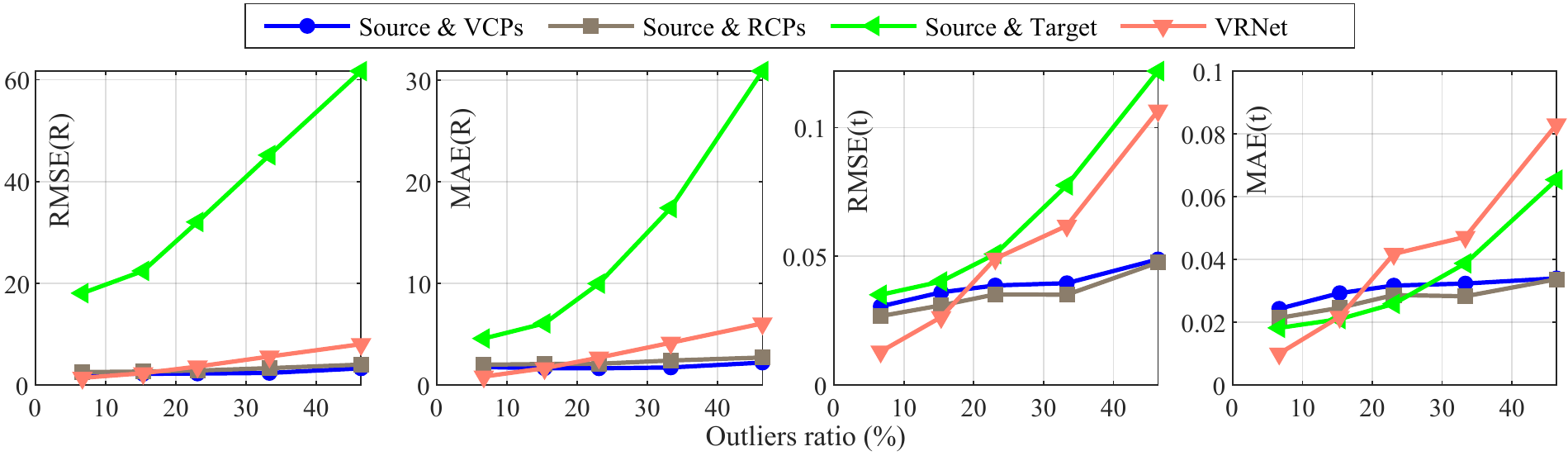}
    \caption{The performance comparison of RANSAC strategy and our proposed correction-walk module with the different outliers. 
    }
    \label{Fig:ransac}
\end{figure*}

\begin{figure}[h]
    \centering
    \includegraphics[width=\linewidth]{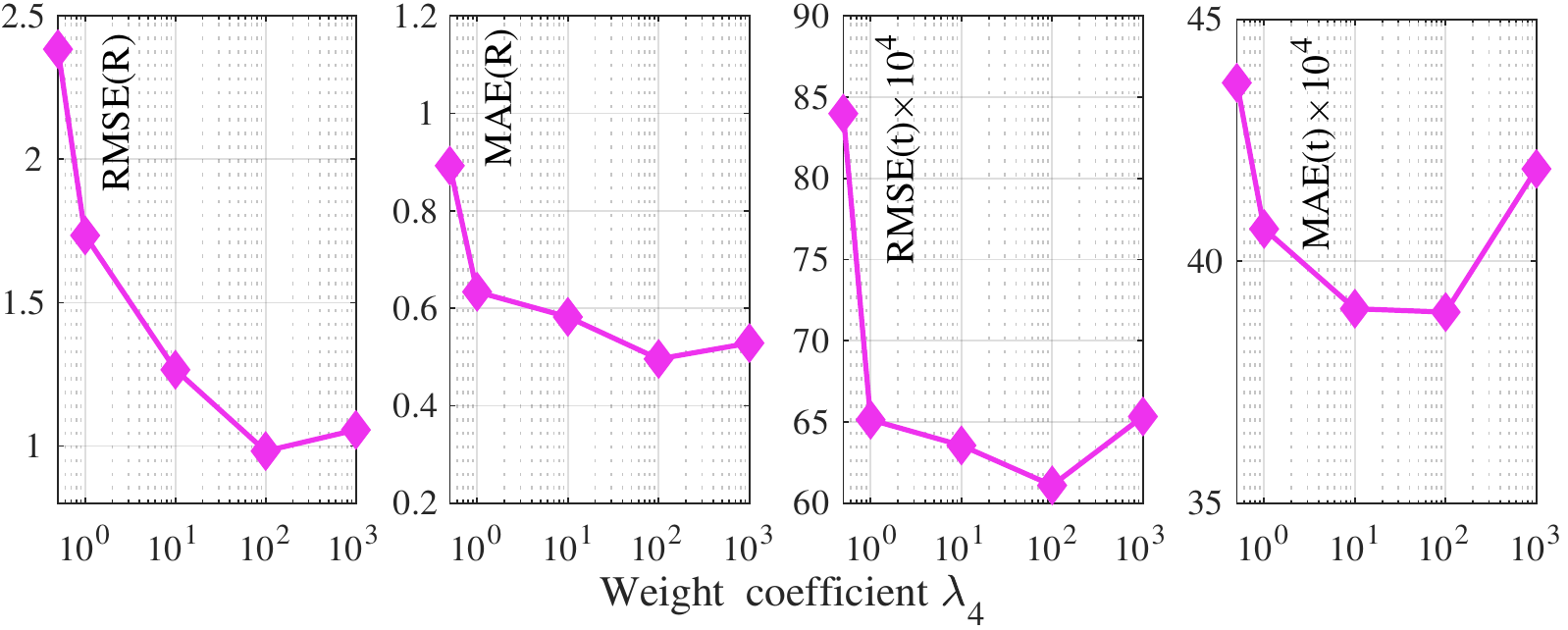}
    \caption{The illustration of registration performance with respect to different weights of the amendment offset supervision, \ie $\lambda_4$. X-axis is labeled with this weight coefficient $\lambda_4$. We notice that, the proposed method achieves the best performance when $\lambda_4 = 100$, which is taken in our applications.}
    \label{Fig:weight}
    \vspace{-0.4cm}
\end{figure}

\noindent\textbf{Coefficients of loss functions.}
In this paper, we take a hybrid loss function for point cloud registration. As mentioned in \secref{3-2-loss}, corresponding point supervision $\mathcal{L}_0$ is used to train the feature extractor and the local motion consensus $\mathcal{L}_1$, geometry structure supervision $\mathcal{L}_2$, $\mathcal{L}_3$ and the amendment offset supervision $\mathcal{L}_4$ are used to train the correction-walk module. Here, $\mathcal{L}_1$, $\mathcal{L}_2$, $\mathcal{L}_3$ are unsupervised, where we take $\lambda_1 = \lambda_2 = \lambda_3 = 1$. And $\mathcal{L}_4$ is supervised and we have a test about $\lambda_4$ in partial-view setting.

As shown in \figref{Fig:weight}, our proposed VRNet achieves the best performance when $\lambda_4 = 100$, which is taken in our application. Besides, we compare the performance of tests conducted with different loss function combinations in \tabref{Tab:weight}. We find that the unsupervised loss functions not only help keep the shape and geometry structure of the learned corresponding points, but also improve the final registration performance.

\begin{table}[h]
	\renewcommand\arraystretch{1.0}
    	\caption{The comparison of different loss function combinations applied in correction-walk module training.}	
    	\vspace{-3mm}
    	\begin{center}
		\resizebox{0.8\linewidth}{!}{
			\begin{tabular}{ccccc}
				\toprule
				\textbf{Methods} &  \textbf{RMSE(R)} & \textbf{MAE(R)} & \makecell[c]{\textbf{RMSE(t)}\\$(\times 10^{-2})$} & \makecell[c]{\textbf{MAE(t)}\\$(\times 10^{-2})$}  \\
				\midrule				
				$\mathbf{\mathcal{L}_4}$                &1.254&0.639&0.838&0.540\\
				$\mathbf{\mathcal{L}_4}$+$\mathbf{\mathcal{L}_1}$&1.062&0.593&0.672&0.403\\
				$\mathbf{\mathcal{L}_4}$+$\mathbf{\mathcal{L}_2}$&1.054&0.539&0.638&0.410\\
				$\mathbf{\mathcal{L}_4}$+$\mathbf{\mathcal{L}_3}$&1.119&0.608&0.715&0.462\\
				%\midrule
				\textbf{All loss}       &\textbf{0.982}&\textbf{0.496}&\textbf{0.611}&\textbf{0.389}\\
				\bottomrule
		\end{tabular}}
		\label{Tab:weight}
	\end{center}
\end{table}

\noindent\textbf{VRNet with iteration.}
Currently, our method has achieved a high-quality registration performance in a single pass. Here, we provide an evaluation to test our VRNet in the iteration strategy like ICP. Specifically, in each iteration, we refine the \textit{source} point clouds with the predicted transformation matrix solved in the last iteration, and resolve the new transformation matrix between the updated \textit{source} point clouds and the \textit{target} point clouds. Finally, all predicted transformation matrices are summarized to obtain the final estimated transformation matrix. The tests are conducted on ModelNet40 under the \textbf{\textit{UPC}} using the \textbf{\textit{partial-view}}. The results are provided in \tabref{Tab:iteration}.

\begin{table}[h]
	\renewcommand\arraystretch{1.0}
    	\caption{The performance comparison when inserting the VRNet to the iteration pipeline.}	
    	\vspace{-3mm}
    	\begin{center}
		\resizebox{0.8\linewidth}{!}{
			\begin{tabular}{cccccc}
				\toprule
				\textbf{Iteration} & \textbf{RMSE(R)} & \textbf{MAE(R)} & \makecell[c]{\textbf{RMSE(t)}\\$(\times 10^{-2})$} & \makecell[c]{\textbf{MAE(t)}\\$(\times 10^{-2})$}&  \textbf{Times}[ms]  \\
				\midrule				
				1& 0.982 &0.496 &0.611 &0.389 & 19.92\\
				2& 0.931 &0.447 &0.563 &0.327 & 41.73\\
				3& 0.904 &0.412 &0.528 &0.301 & 65.34\\
				4& 0.886 &\textbf{0.398} &0.501 &0.292 & 96.42\\
				5& \textbf{0.884} &0.401 &0.497 &0.292 & 127.51\\
				6& 0.891 &0.399 &\textbf{0.495} &\textbf{0.291} & 162.46\\
				\bottomrule
		\end{tabular}}
		\label{Tab:iteration}
	\end{center}
 	\vspace{-0.5cm}
\end{table}

From \tabref{Tab:iteration}, we can find that, before the $4$-th iteration, the registration performance has been significantly improved as the number of iterations increases. However, when the number of iterations increases further, the performance tends to be stable. The time-efficiency becomes weaker due to the iteration strategy.

\noindent\textbf{Robustness to outliers.}
To verify the robustness to outliers, we evaluate the registration performance of ours and the baselines under different outliers ratios as shown in \figref{Fig:degeneration}. The tests are conducted on ModelNet40 under the \textbf{\textit{UPC}} using the \textbf{\textit{partial-view}}. By different sample ratios, different outliers ratios are achieved. And the fewer points are sampled, the higher outliers ratio are obtained. Specifically, there are 1024 points in the each original consistent input point cloud, whether it is the \textit{source} or the \textit{target}. We reconstruct the \textit{source} and the \textit{target} by sampling some nearest points from them. The corresponding relationship between the size of the sampled point cloud and the outliers ratio is 960 (6.67\%), 896 (15.33\%), 832 (23.08\%), 768 (33.33\%), 704 (46.39\%).
From \figref{Fig:degeneration}, obviously, for rotation estimation, VRNet achieves stable and the most accurate performance. For translation estimation, FGR achieves the better performance. However, for learning-based methods, ours remains the best performance\footnote{In the ablation study of ``robustness to outliers'', ``correct matches ratio vs outliers ratio'', ``RANSAC vs correction-walk'' and ``visualization'', for a clear comparison and analysis of the proposed VRNet, we adjust the \textbf{\textit{partial-view}} setting proposed in PRNet\cite{wang_prnet_nips_2019}. Specifically, we set the view points at the symmetrical position to sample the \textit{source} and \textit{target} rather than the same position. This operation will result in a larger outliers ratio and more obvious partiality. These tests are conducted on ModelNet40 under \textbf{\textit{UPC}}.}.

\begin{figure}[h]
    \begin{center}
    \includegraphics[width=0.9\linewidth]{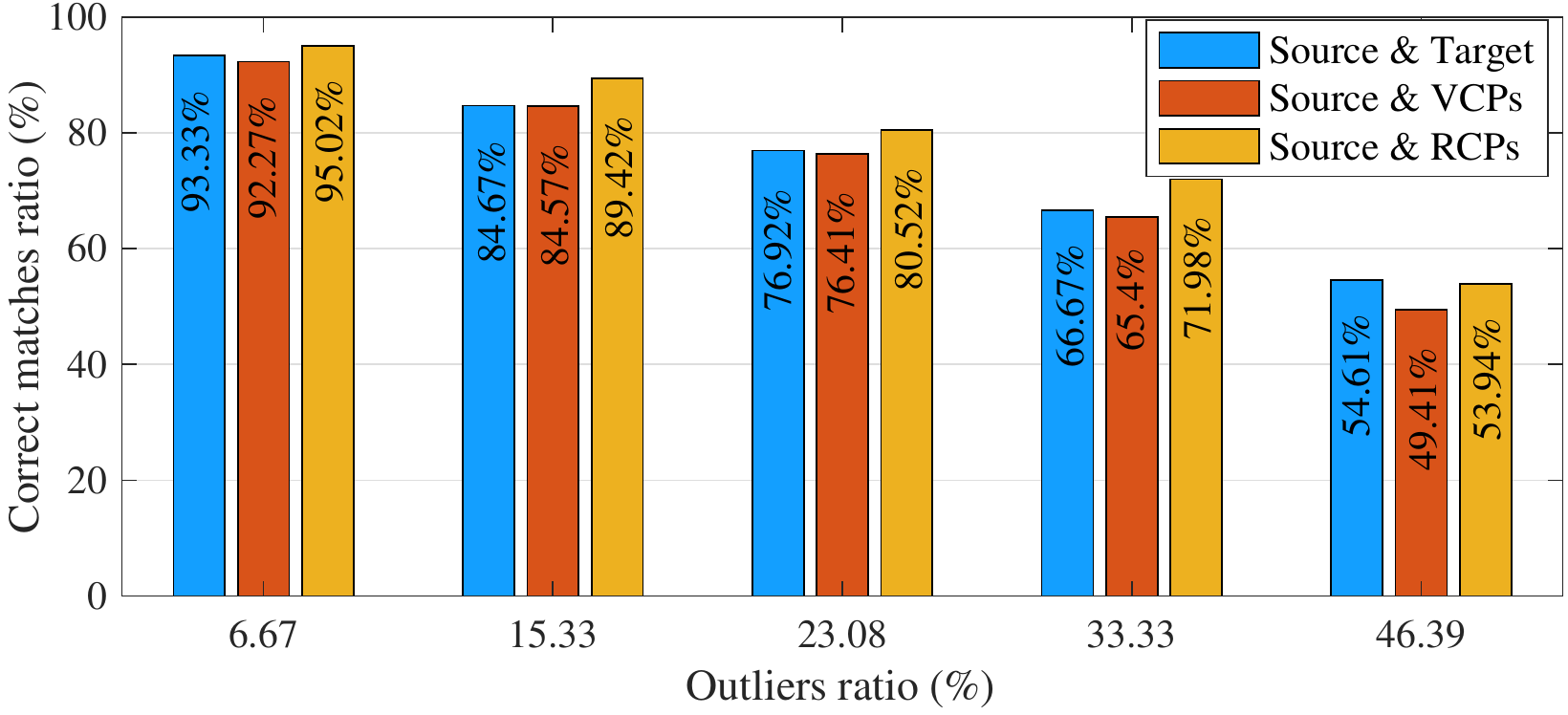}
    \vspace{-0.2cm}
    \caption{The correct matches ratio comparison with the different outliers ratios. }
    \vspace{-0.5cm}
    \label{Fig:matches}
    \end{center}
\end{figure}

\noindent\textbf{Correct matches ratio vs. outliers ratio.}
We evaluate the correct matches ratio as the outliers ratio increases in \figref{Fig:matches}. 
We compare the ground truth, \ie the correct matches ratio of the \textit{source} point cloud and the \textit{target} point cloud, the correct matches ratio of the \textit{source} and the learned VCPs, and the correct matches ratio of the \textit{source} and the learned RCPs. The thresholds to confirm the successful matches are set to 0.15, \ie, if the distance between the predicted corresponding point and the ground truth corresponding point is less than the threshold, this match is a successful match.
From \figref{Fig:matches}, the correct matches ratio of the \textit{source} and the VCPs approximates the ground truth, which validates our method leaned an accurate corresponding points distribution of inliers due to the proposed sufficient supervision. And the correct matches ratio of the \textit{source} and RCPs is even better than the ground truth, which verifies that the distribution of the learned RCPs is more 
consistent with the \textit{source} than the original \textit{target} point cloud due to the correction-walk module.

\begin{figure*}[!h]
    \centering
    \includegraphics[width=0.9\textwidth]{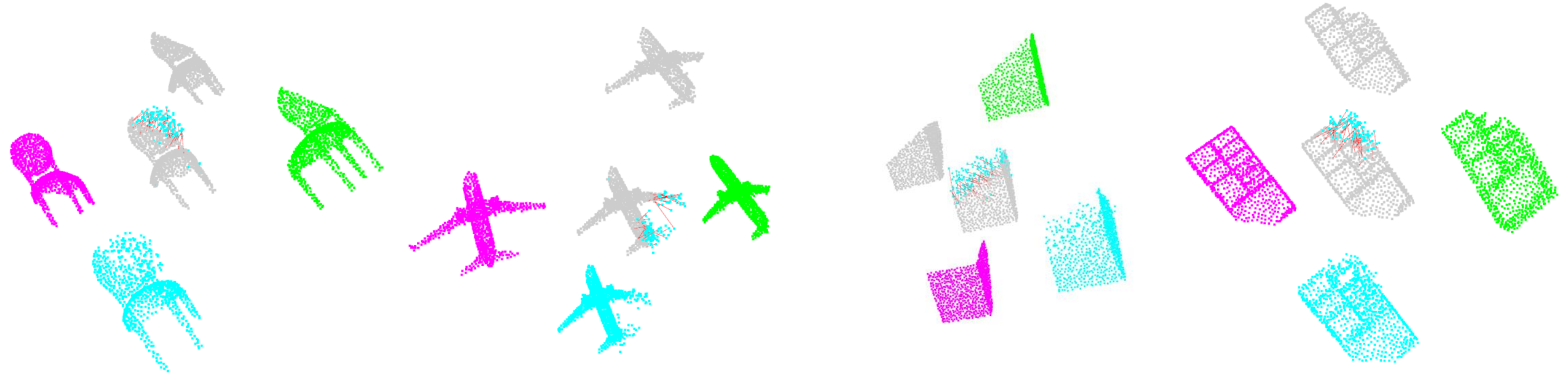}
    \vspace{-0.2cm}
    \caption{Visualization of the \textit{source} point (purple), the \textit{target} point cloud (green), the learned VCPs (gray), RCPs (blue), and the learned offset (red line). All points clouds are calibrated to the same pose for clear comparison. Obviously, the VCPs approximate the \textit{source} as much as possible but it is limited in the \textit{target} distribution. Then, the correction-walk module amends the VCPs to the RCPs, which present a more consistent distribution with the \textit{source} than the VCPs and the origianl \textit{target}.}
    \label{Fig:vis}
\end{figure*}

\begin{figure*}[!ht]
    \centering
    \vspace{-0.4cm}
    \includegraphics[width=1.0\textwidth]{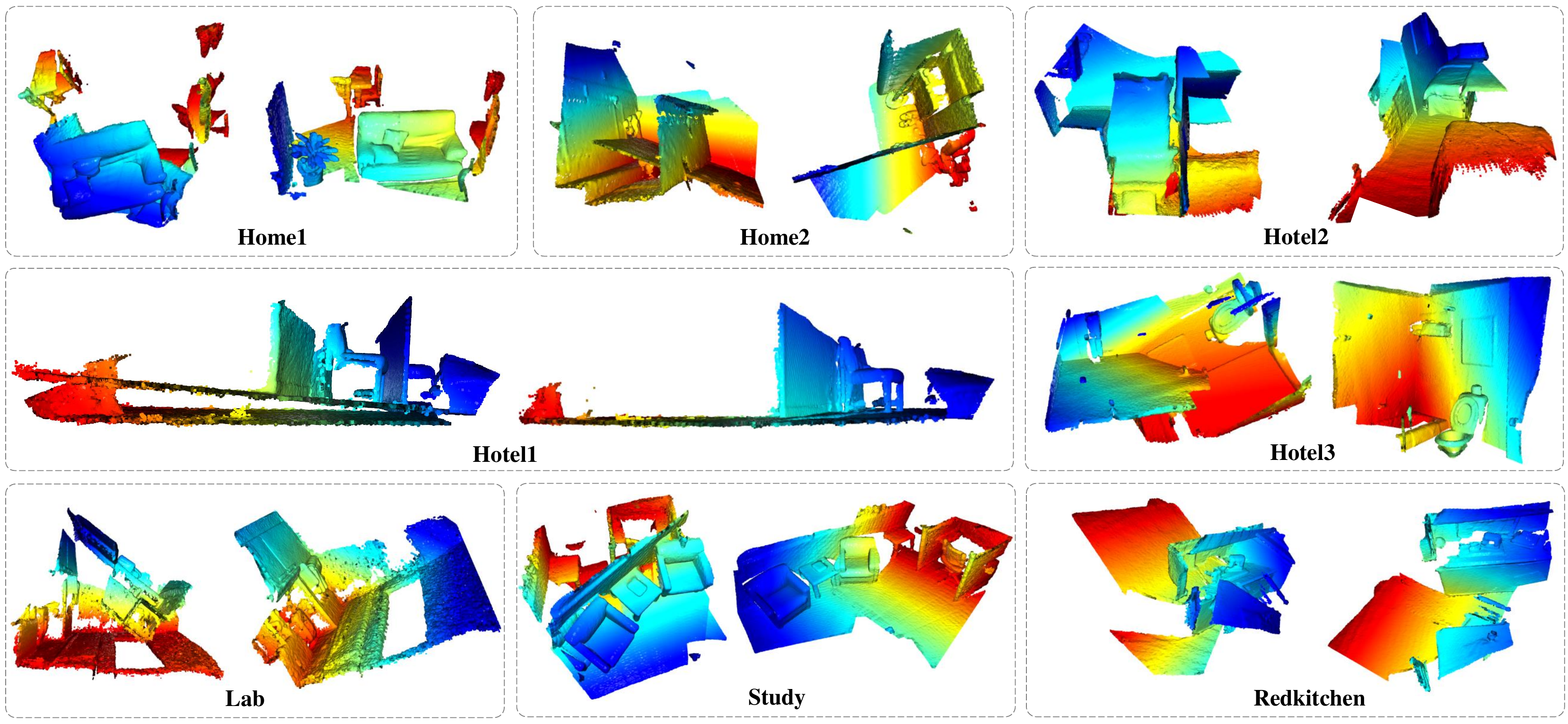}
    \vspace{-0.4cm}
    \caption{Our VRNet presents the accurate registration results on real indoor dataset, 3DMatch including 8 subsets, where \textbf{Left} is the input \textit{source} and \textit{target} point clouds, and \textbf{Right} is the aligned point clouds.}
    \label{Fig:vis_3dmatch}
\end{figure*}

\begin{figure*}[!ht]
    \centering
    \vspace{-0.4cm}
    \includegraphics[width=0.9\textwidth]{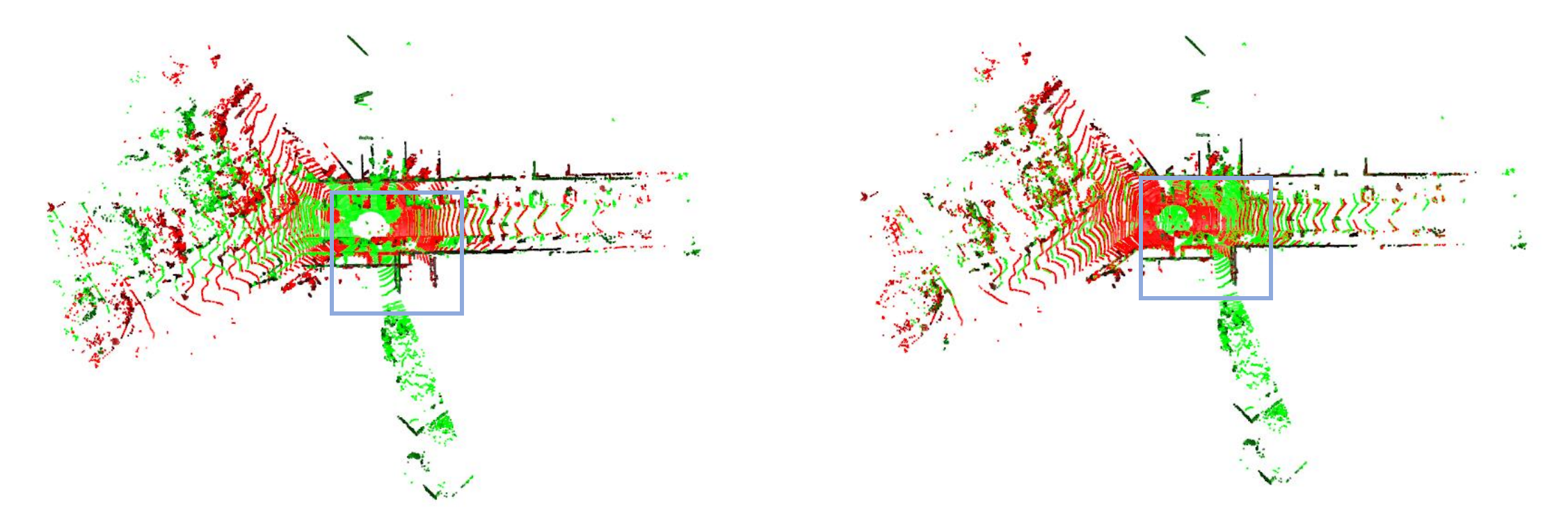}
    \vspace{-0.2cm}
    \caption{Our VRNet achieves accurate registration performance on the real outdoor KITTI dataset. The left subfigure is the input pair and the right subfigure indicates the registration result of our VRNet. We can find the input \textit{source} and \textit{target} point clouds are with different poses while the point clouds in the right subfigure are aligned. For example, in the left subfigure, the edges of the road are biased. And the blank parts of the center overlap marked by the blue box but they should be different in fact because the scanner locates at different positions. Then, in the right subfigure, the edges overlap, and the blank parts deviate since the point clouds have been registered successfully.}
    \label{Fig:vis_kitti}
\end{figure*}

\noindent\textbf{RANSAC vs. correction-walk.}
The principle behind the VRNet is to consider the \textit{source} uniformly without distinguishing inliers and outliers rather than selecting reliable correspondences. Here, we evaluate these two strategies. RANSAC, which is the representative correspondences selection method, is compared here. The results of RANSAC, which are applied to \textit{source} point cloud and the \textit{target} point cloud, are presented in \figref{Fig:ransac}. Obviously, our VRNet achieves more accurate results with different outliers ratios than RANSAC in terms of rotation estimation and RMSE(t). With respect to MAE(t), when the outliers ratio is high, RANSAC method is better while ours obtains more accurate results when the outliers ratio is low. To sum up, VRNet achieves better performance than RANSAC, especially facing low outliers ratio.
Besides, we test the performance of combining the RANSAC and our VRNet. Two settings are carried out here, \ie employing RANSAC to the \textit{source} and the learned VCPs, employing RANSAC to the \textit{source} and the learned RCPs. 
From \figref{Fig:ransac}, we can find that, RANSAC between the \textit{source} and the learned RCPs gets a litter better performance than RANSAC between the \textit{source} and the learned VCPs. Meanwhile, these two settings are both much more stable than using RANSAC or VRNet alone when outliers ratio increases. It is worth mentioning that, when outliers ratio is low, especially approximates to 0, RANSAC provides negative affect to our VRNet. We suspect that the tolerance in RANSAC strategy (\ie the threshold to determine the reliable correspondences) decreases the registration performance.

\noindent\textbf{Visualization}
We provide the visualization of the learned VCPs, the learned RCPs and the offset learned by the correction-walk module in \figref{Fig:vis}. VCPs are limited in the \textit{target} point cloud, however, they are amended to RCPs that are more consistent with the \textit{source} point cloud. In addition, for a clear presentation of the effectiveness of the proposed VRNet, we provide some registration results on 3DMatch and KITTI datasets in \figref{Fig:vis_3dmatch} and \figref{Fig:vis_kitti}.

%%%%%%%%%%%%%%%%%%%%%%%%%%%%%%%%%%%%%%%%%%%%%%%%%
\section{Discussion and conclusion} \label{5_conclusion}
%%%%%%%%%%%%%%%%%%%%%%%%%%%%%%%%%%%%%%%%%%%%%%%%%
In this paper, we have proposed VRNet, an end-to-end robust 3D point cloud registration network. However, some limitations also exist in our method. Specifically, 1) our proposed method cannot handle the object or scene with strong symmetry well. Our method advocates learning the correction displacement by comparing the features of the \textit{source} point clouds and the virtual point clouds to facilitate the RCPs to tend to be consistent with the \textit{source} point clouds. However, if the point features are confused due to the same geometry in the symmetric object, the learned offsets are also confused, which can not rectify the VCPs accurately; 
2) even though our method has presented strong robustness as the overlap ratio decreases, when the overlap ratio becomes very low, the registration performance also will be significantly affected. This is a stubborn illness in the point cloud registration field \cite{wang_dcp_iccv_2019,wang_prnet_nips_2019,choy_dgr_cvpr_2020}. For our method, we suspect the reason is that the learned RCPs cannot be built accurately since there are too many outliers in the \textit{source} that need to be fitted by the virtual points.

Nevertheless, our VRNet can effectively avoid the complicated inliers and outliers screening and reliable correspondences selection by modeling the corresponding points for all \textit{source} points uniformly.
Its is proven to be effective and efficient in recovering the rectified virtual corresponding points that maintain the same shape as the \textit{source} and same pose of the \textit{target}, thanks to the use of the proposed correction-walk module and the hybrid loss function. 
Our experiments show that VRNet can achieve state-of-the-art rigid transformation estimation results and high time-efficiency on both synthetic and real sparse datasets.
Meanwhile, for large-scale dense datasets, VRNet can balance time-efficiency and accuracy. It not only achieves comparable performance as the most advanced methods but also maintains good time superiority, which is crucial for practical applications.
In the future, we plan to extend our VRNet to 2D-2D and 2D-3D registrations. 
In addition, we would further investigate more effective downsampling strategies to help VRNet improve the registration accuracy for large-scale dense point cloud data.

%%%%%%%%%%%%%%
\section*{Acknowledgement}
\thanks{This work was supported in part by the National Key Research and Development Program of China under Grant 2018AAA0102803 and National Natural Science Foundation of China (61871325, 61901387, 62001394). This work was also sponsored by Innovation Foundation for Doctor Dissertation of Northwestern Polytechnical University.}

%%%%%%%%%%%%%%Reference
%\bibliographystyle{unsrt}
%\bibliographystyle{alpha}
%\bibliography{VRNetBib}

\end{document}